\documentclass[final]{cvpr} 

\usepackage{lmodern, times, epsfig, graphicx, amsmath, amssymb}
\usepackage{xcolor, tabulary, xfrac, enumitem, setspace}
\usepackage[british, english, american]{babel}
\usepackage[pagebackref=true, breaklinks=true, letterpaper=true, colorlinks,
            citecolor=citecolor, bookmarks=false]{hyperref}
\definecolor{citecolor}{RGB}{34,139,34}

\def\confYear{CVPR 2021}

\renewcommand{\paragraph}[1]{\vspace{1.25mm}\noindent\textbf{#1}}
\newcommand{\eqnsm}[3]{\vspace{#1}\begin{equation}\label{eq:#2}#3\vspace{#1}\end{equation}\ignorespaces}
\newcommand{\tablestyle}[2]{\setlength{\tabcolsep}{#1}\renewcommand{\arraystretch}{#2}\centering\footnotesize}
\newlength\savewidth\newcommand\shline{\noalign{\global\savewidth\arrayrulewidth
  \global\arrayrulewidth 1pt}\hline\noalign{\global\arrayrulewidth\savewidth}}

\newcommand{\tightboxed}[1]{\setlength{\fboxsep}{.3\fboxsep}\boxed{#1}}

\newcommand{\app}{\raise.17ex\hbox{$\scriptstyle\sim$}}
\newcommand{\mypm}[1]{\color{gray}{\tiny{$\pm$#1}}}
\newcommand{\x}{{\times}}

\definecolor{scolor}{RGB}{30,60,180}
\newcommand{\s}{{\color{scolor}s}}
\renewcommand{\ss}{{\color{scolor}\sqrt{s}}}
\newcommand{\sss}[1]{{\color{scolor}\sqrt[#1]{s}}}
\newcommand{\spow}[1]{{\color{scolor}s^{#1}}}

\begin{document}
\title{Fast and Accurate Model Scaling\\[-3mm]}
\author{%
 Piotr Doll\'ar \quad Mannat Singh \quad Ross Girshick\\[2mm]
 Facebook AI Research (FAIR)\vspace{-1mm}}
\maketitle

\begin{abstract}
In this work we analyze strategies for convolutional neural network scaling; that is, the process of scaling a base convolutional network to endow it with greater computational complexity and consequently representational power. Example scaling strategies may include increasing model width, depth, resolution, \etc. While various scaling strategies exist, their tradeoffs are not fully understood. Existing analysis typically focuses on the interplay of accuracy and flops (floating point operations). Yet, as we demonstrate, various scaling strategies affect model parameters, activations, and consequently actual runtime quite differently. In our experiments we show the surprising result that numerous scaling strategies yield networks with similar accuracy but with widely varying properties. This leads us to propose a simple \emph{fast compound scaling} strategy that encourages primarily scaling model width, while scaling depth and resolution to a lesser extent. Unlike currently popular scaling strategies, which result in about $O(s)$ increase in model activation \wrt scaling flops by a factor of $s$, the proposed fast compound scaling results in close to $O(\sqrt{s})$ increase in activations, while achieving excellent accuracy. Fewer activations leads to speedups on modern memory-bandwidth limited hardware (\eg, GPUs). More generally, we hope this work provides a framework for analyzing scaling strategies under various computational constraints.
\end{abstract}\vspace{-2mm}

\section{Introduction}

\begin{figure}[t]\centering
\includegraphics[height=2.58cm]{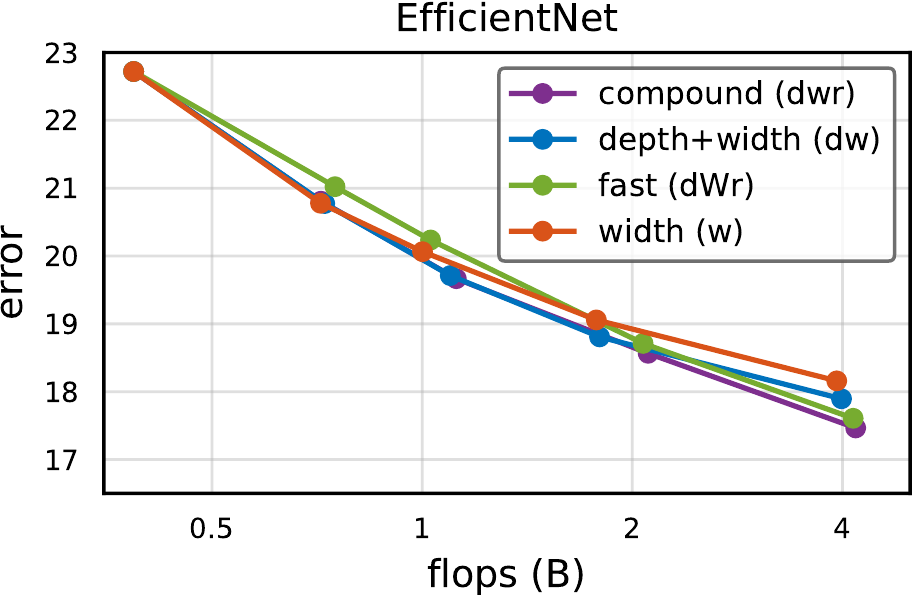}\hspace{2mm}
\includegraphics[height=2.58cm]{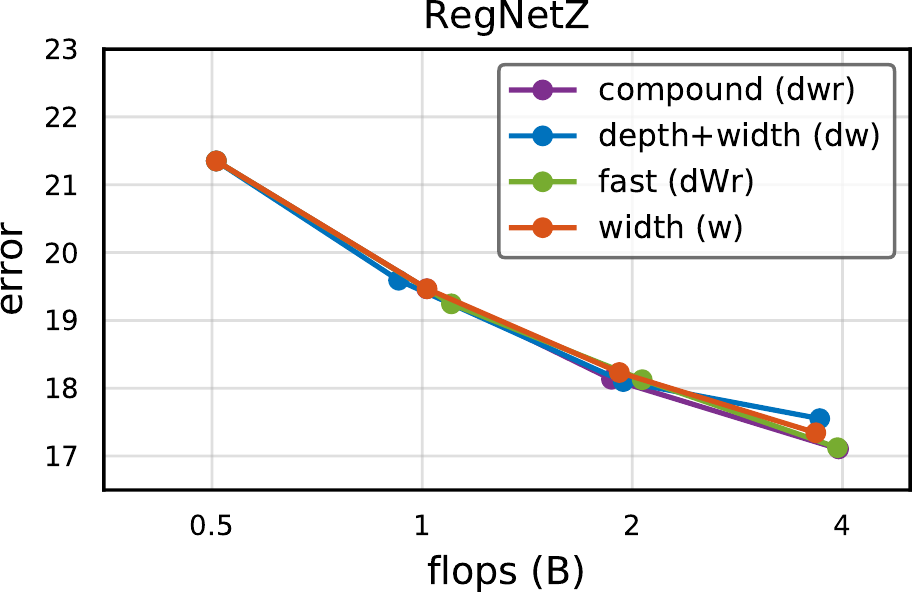}\\[1mm]
\includegraphics[height=2.58cm]{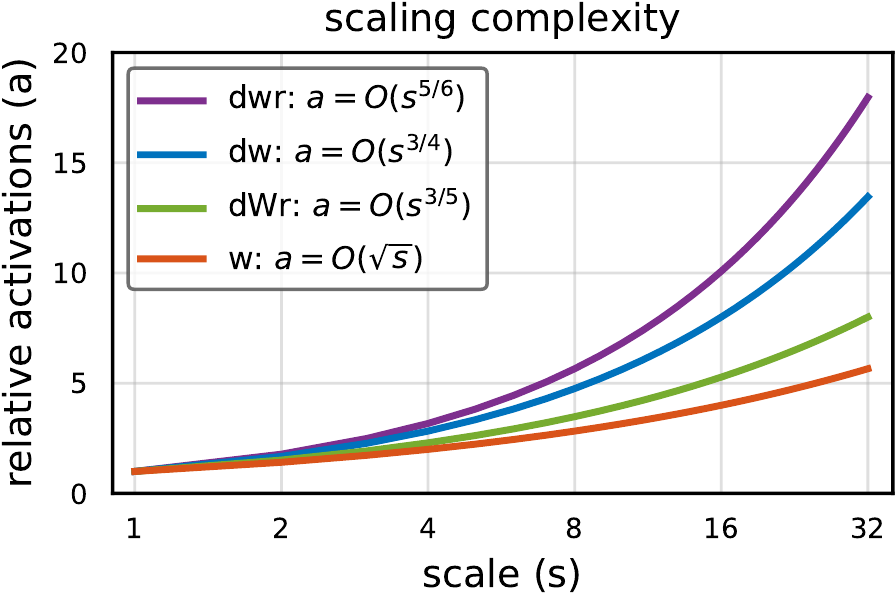}\hspace{2mm}
\includegraphics[height=2.58cm]{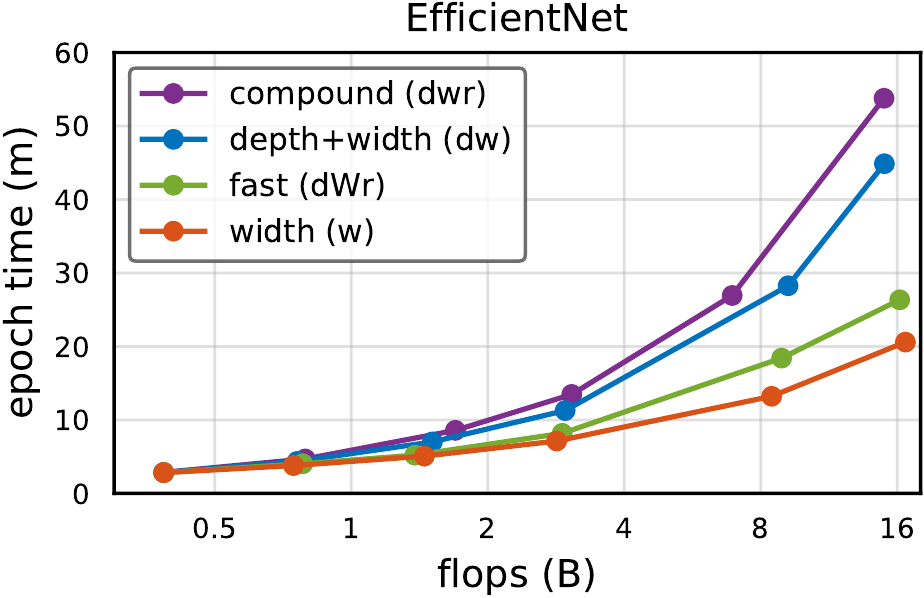}\\
\caption{An analysis of four model scaling strategies: \emph{width} scaling ($w$), in which only the width of a base model is scaled; \emph{compound} scaling ($dwr$), in which the width, depth, and resolution are all scaled in roughly equal proportions; \emph{depth and width} scaling ($dw$); and the proposed \emph{fast compound} scaling ($dWr$), which emphasizes scaling primarily, but not only, the model width. \emph{(Top):} We apply the four scaling strategies to two base models (EfficientNet-B0 and RegNetZ-500MF). Compound and fast scaling result in highest accuracy models, and both outperform width scaling. \emph{(Bottom-left):} The scaling strategies have asymptotically different behavior in how they affect model activations. Given a scale factor of $s$, activations increase with about $O(\sqrt{s})$ for $w$ and $dWr$ scaling compared to almost $O(s)$ for $dwr$ and $dw$ scaling. \emph{(Bottom-right):} Runtime of a model (EfficientNet-B0) scaled using the four scaling strategies. Fast scaling results in models nearly as fast as $w$ scaling (but with higher accuracy), and much faster than $dwr$ and $dw$ scaling, closely reflecting model activations.}
\label{fig:teaser}\vspace{-2mm}
\end{figure}

Advances in modern hardware for training and running convolutional neural networks over the past several years have been impressive. Highly-parallel hardware accelerators, such as GPUs and TPUs, allow for training and deploying ever larger and more accurate networks.

Interestingly, this rapid advancement has greatly benefited our ability to optimize models for the \emph{low-compute} regime. In particular, whether via manual design, random search, or more complex neural architecture search strategies~\cite{Zoph2017}, it has become feasible to train a large number of small models and select the best one, in terms of both accuracy \emph{and} speed. At intermediate-compute regimes, efficient search~\cite{Liu2018} or efficient design spaces~\cite{Radosavovic2019, Radosavovic2020} can still provide the ability to directly optimize neural networks. However, regardless of computational resources, there will necessarily exist a \emph{high-compute} regime where it may only be feasible to train a handful of models, or possibly even only a \emph{single} model. This regime motivates our work.

In the high-compute regime, \emph{network scaling}, the process by which a lower-complexity model is enlarged by expanding one or more of its dimensions (\eg, depth or width), becomes essential. Scaling has proven effective in terms of obtaining larger models with good accuracy~\cite{Tan2019}. However, existing work on model scaling focuses on model \emph{accuracy}. In this work, we are interested in large, accurate models that are \emph{fast} enough to deploy and use in practice.

The concept of network scaling emerged naturally in deep learning, with early work focused on scaling networks by increasing depth~\cite{Simonyan2015, Szegedy2015, He2016}. However, gains from depth scaling plateaued, leading to explorations of scaling width~\cite{Zagoruyko2016} and resolution~\cite{Howard2017}. More recently scaling multiple dimensions at once, coined \emph{compound scaling}~\cite{Tan2019}, has been shown to achieve excellent accuracy.

Existing explorations of model scaling typically focus on maximizing \emph{accuracy} versus \emph{flops}. Yet as we will show, two scaled models with the same flops can have very different runtime on modern accelerators. This leads us to the central question explored in our work: \emph{can we design scaling strategies that optimize both accuracy and model runtime?}

Our first core observation is that there exists multiple scaling strategies that can yield similar accuracy models at the same flops. In Figure~\ref{fig:teaser}, top, we show that there exist multiple scaling strategies that can result in models with high accuracy. We will expand on this result in \S\ref{sec:exps}.

However, scaling a model to a fixed target flops using two scaling strategies can result in widely different runtimes, see Figure~\ref{fig:teaser}, bottom-right. To better understand this behavior at a more fundamental level, in \S\ref{sec:complexity} we develop a framework for analyzing the complexity of various scaling strategies, in terms of not just flops, but also parameters and activations. In particular, we show that different strategies scale activations at different asymptotic rates relative to flops. \Eg, when scaling a model from $f$ flops to $sf$ flops by scaling width, activations increase by $O(\sqrt{s})$, compared to nearly $O(s)$ for compound scaling. Figure~\ref{fig:teaser}, bottom-left, shows this asymptotic behavior for a few select strategies. 

In \S\ref{sec:runtime} we will show that within a flop range of practical interest, on modern accelerators \emph{the runtime of a scaled model is more strongly correlated with activations than flops}. We emphasize that this correlation holds over a diverse set of scaling strategies, which enables us to use activations as a proxy for predicting a scaled model's runtime.

Based on our analysis, in \S\ref{sec:fast} we introduce a new family of scaling strategies parameterized by a single parameter $\alpha$ that controls the relative scaling along model width versus other dimensions. This lets us carefully control the asymptotic rate at which model activations scale. We show $0\ll\alpha<1$ yields models that are both fast and accurate. We refer to this scaling strategy as fast compound model scaling, or simply \emph{fast scaling} for brevity. 

As we will show in \S\ref{sec:exps}, fast scaling allows us to obtain large models that are as accurate as the state-of-the-art but faster. As a concrete example, we apply fast scaling to scale a RegNetY-4GF~\cite{Radosavovic2020} model to 16GF (gigaflops), and find it uses less memory and is faster (and more accurate) than EfficientNet-B4~\cite{Tan2019} -- a model with 4$\times$ fewer flops.

In order to facilitate future research we will release all code and pretrained models introduced in this work.\footnote{\url{https://github.com/facebookresearch/pycls}}

\section{Related Work}\vspace{-2mm}

\paragraph{Manual network design.} Since the impressive success of AlexNet~\cite{Krizhevsky2012}, and with the steady progress of hardware accelerators, the community has pushed toward ever larger and more accurate models. Increasing model depth led to rapid gains, notable examples include VGG~\cite{Simonyan2015} and Inception~\cite{Szegedy2015, Szegedy2016a}. This trend culminated with the introduction of residual networks~\cite{He2016}. Next, wider models proved not only effective but particularly efficient~\cite{Zagoruyko2016, Howard2017}. The use of depthwise~\cite{Chollet2017} and group convolution~\cite{Xie2017} enabled even higher capacity models. Other notable design elements that led to larger and more accurate models include the inverted bottleneck~\cite{Sandler2018}, SE~\cite{Hu2018}, and new nonlinearities~\cite{Hendrycks2016, Ramachandran2017}.

\paragraph{Automated network design.} With the rapid advancement of hardware for training deep models, it has become more feasible to \emph{automate} network design. Neural architecture search~\cite{Zoph2017, Zoph2018, Real2018} has turned into a thriving research area and led to highly efficient models, especially in the low-compute regime. Model search is computationally expensive when training larger models, this has led to interest in developing efficient search algorithms~\cite{Liu2018, Pham2018, Liu2019}. For example, DARTS~\cite{Liu2019} proposed a differentiable search strategy that does not require training multiple separate models to optimize model structure. Nevertheless, in practice search is most effective in low or medium compute regimes.

\paragraph{Design space design.} Despite the effectiveness of model search, the paradigm has limitations. The outcome of a search is a single model \emph{instance} tuned to a specific setting (\eg, dataset or flop regime). As an alternative, Radosavovic \etal~\cite{Radosavovic2020} recently introduced the idea of \emph{designing design spaces}, and designed a low-dimensional design space consisting of simple, easy-to-tune models. Given a new dataset or compute regime, a model can be selected from this design space by tuning a handful of parameters, allowing for highly efficient random search. This allows for optimizing models directly in fairly high-compute regimes. We utilize these efficient design spaces in our experiments.

\paragraph{Network scaling.} Regardless of the model design strategy, there will exist some computational regime in which it is not feasible to train and compare a large number of models. Thus model scaling becomes crucial. Popular scaling strategies include scaling depth~\cite{Simonyan2015, Szegedy2015, He2016}, width~\cite{Zagoruyko2016, Howard2017} and resolution~\cite{Howard2017, Huang2019gpipe}. The recently introduced compound scaling strategy~\cite{Tan2019b}, which scales along all three dimensions at once, achieves an excellent accuracy versus flops tradeoff and serves as a core baseline in our work.
 
\paragraph{Going bigger.} There is substantial interest in scaling to massive datasets~\cite{Sun2017jft, Mahajan2018} and compute regimes~\cite{Huang2019gpipe}. Moreover, recent progress in unsupervised learning~\cite{He2020moco, Chen2020simclr, Caron2020swav} may create the potential to train with essentially unlimited data. These efforts motivate our work: we aim to enable scaling models to the size necessary for these brave new regimes.

\section{Complexity of Scaled Models}\label{sec:complexity}

\begin{table}\centering\vspace{-1mm}
\tablestyle{4.5pt}{1.00}
\begin{tabular}{@{}c|rrr|rrr@{}}
 dim & \multicolumn{3}{c|}{scaling} & flops $(f)$ & params $(p)$ & acts $(a)$\\\shline
 none & $d$ & $w$ & $r$ & $d w^2 r^2$ & $d w^2$ & $d w r^2$ \\
 $d$ & $\s d$ & $w$ & $r$ & $\s d w^2 r^2$ & $\s d w^2$ & $\s d w r^2$ \\
 $w$ & $d$ & $\ss w$ & $r$ & $\s d w^2 r^2$ & $\s d w^2$ & $\tightboxed{\ss d w r^2}$ \\
 $r$ & $d$ & $w$ & $\ss r$ & $\s d w^2 r^2$ & $\tightboxed{{\color{scolor}1} d w^2}$ & $\s d w r^2$ \\
\end{tabular}\vspace{1.5mm}
\caption{\textbf{Simple scaling}: Complexity of scaling a stage (of $d$ conv layers with width $w$ and spatial resolution $r \x r$) by a factor of $s$ using various simple scaling strategies where a single dimension is varied at a time. Note that parameters and activations vary substantially for the scaling strategies, especially for large $s$.}
\label{tab:scaling:single}\vspace{-1mm}
\end{table}

In this section we present a general framework for analyzing the complexity of various network scaling strategies. While the framework is simple and intuitive, it proves powerful in understanding and extending model scaling.

\subsection{Complexity Metrics}\label{sec:complexity:metrics}

The three most relevant properties of models we consider are their \emph{flops ($f$)}, \emph{parameters ($p$)}, and \emph{activations ($a$)}. Following common practice, we use flops to mean multiply-adds and parameters to denote the number of free variables in a model. We define activations as the number of elements in the output tensors of convolutional (conv) layers. 

Flops and parameters are popular complexity measures of neural networks. We note, however, that parameters of a convolution are independent of input resolution and hence do not fully reflect the actual capacity or runtime of a convolutional network. Therefore, given that we study networks with varying input resolution, we report parameters but \emph{we focus on flops as a primary complexity measure}.

Activations are less often reported but as we demonstrate play a key role in determining network speed on modern memory-bandwidth limited hardware. Hence, we carefully analyze the interplay between scaling and activations. 

\subsection{Network Complexity}\label{sec:complexity:network}

While conv networks are composed of many heterogeneous layers, we focus our complexity analysis on conv layers. First, many layers such as normalization, pooling, or activation often account for a small percentage of a model's compute. Second, the number and complexity of these layers tends to be \emph{proportional} to the number and size of conv layers (\eg, every conv may be followed by an activation). For these reasons analyzing convs serves as an excellent proxy of how model scaling affects an entire network.

Consider a $k \x k$ conv layer with width (number of channels) $w$ and spatial resolution $r$. The layer takes in a feature map of size $r \x r \x w$, and for each of the $r^2$ patches of size $k \x k \x w$ the network applies $w$ dot products of size $wk^2$. Therefore the complexity of a conv layer is given by:
 \eqnsm{-1.5mm}{conv_1}{f = w^2r^2k^2, \quad p = k^2w^2, \quad a = wr^2}
As $k$ is not scaled, we let $k=1$ without loss of generality.

Common networks are composed of \emph{stages}, where each stage consists of $d$ uniform conv layers, each with the same $w$ and $r$. The complexity of a stage of depth $d$ is:
 \eqnsm{-1.5mm}{conv_d}{f = dw^2r^2, \quad p = dw^2, \quad a = dwr^2}
In subsequent analysis we will show how different scaling strategies affect the complexity of a single stage. For simplicity, we use the same scaling for each network stage, thus our complexity analysis applies to the entire network.
 
\begin{table}\centering\vspace{-1mm}
\tablestyle{4.5pt}{1.05}
\begin{tabular}{@{}c|rrr|rrr@{}}
 dims & \multicolumn{3}{c|}{scaling} & flops $(f)$ & params $(p)$ & acts $(a)$\\\shline
 $dw$ & $\ss d$ & $\sss{4}w$ & $r$ & $\s d w^2 r^2$ & $\s d w^2$ & $\spow{3/4} d w r^2$ \\
 $wr$ & $d$ & $\sss{4}w$ & $\sss{4}r$ & $\s d w^2 r^2$ & $\ss d w^2$ & $\spow{3/4} d w r^2$ \\
 $dr$ & $\ss d$ & $w$ & $\sss{4}r$ & $\s d w^2 r^2$ & $\ss d w^2$ & $\s d w r^2$ \\
 $dwr$ & $\sss{3}d$ & $\sss{6}w$ & $\sss{6}r$ & $\s d w^2 r^2$ & $\spow{2/3} d w^2$ & $\spow{5/6} d w r^2$ \\
\end{tabular}\vspace{1.5mm}
\caption{\textbf{Compound scaling}: Complexity of compound scaling strategies with \emph{uniform} scaling along each dimension, where the relative flops increase of scaling along each dimension is equal. Scaling uniformly along all dimensions, which closely resembles compound scaling~\cite{Tan2019}, results in near linear scaling of activations.}
\label{tab:scaling:compound}\vspace{-1mm}
\end{table}

\subsection{Complexity of Simple Scaling}\label{sec:complexity:simple}

We define \emph{simple scaling} of a stage as scaling a stage along a \emph{single} dimension. In particular, we consider width ($w$), depth ($d$), and resolution ($r$) scaling. In addition to the \emph{scaling dimension}, we define the \emph{scaling factor} $s$ to be the amount by which scaling increases model \emph{flops}. Increasing $d$ by $s$, $w$ by $\sqrt{s}$, or $r$ by $\sqrt{s}$ all increase flops by $s$ (for simplicity we ignore quantization effects). 

Table~\ref{tab:scaling:single} shows the complexity of scaling a stage by a factor of $s$ along different scaling dimensions. While in each case the resulting flops are the same (by design), the parameters and activations vary. In particular, activations increase by $\sqrt{s}$ when scaling width compared to by $s$ when scaling along resolution or depth. This observation will play a central role in how we design new scaling strategies.

\subsection{Complexity of Compound Scaling}\label{sec:complexity:compound}

Rather than scaling along a single dimension, an intuitive approach is to scale along multiple dimensions at once. Coined \emph{compound scaling} by~\cite{Tan2019}, such an approach has been shown to achieve higher accuracy than simple scaling.

In Table~\ref{tab:scaling:compound} we show the complexity for scaling along either two or three dimensions. In each case, we select ratios such that scaling is \emph{uniform} \wrt flops along each dimension. \Eg, if scaling along all dimensions ($dwr$), we scale $d$ by $\sqrt[3]{s}$, $w$ by $\sqrt[6]{s}$, and $r$ by $\sqrt[6]{s}$, such that flops increase by $\sqrt[3]{s}$ when scaling each dimension and by $\sqrt[3]{s}^3 = s$ in total.

Interestingly, the compound scaling rule discovered empirically in~\cite{Tan2019} scaled by 1.2, 1.1, and 1.15 along $d$, $w$, and $r$, which corresponds roughly to \emph{uniform compound scaling} with $s=2$ ($\sqrt[3]{s} \approx 1.26$, $\sqrt[6]{s} \approx 1.12$). We thus use uniform compound scaling as a simple proxy for the purpose of our analysis. Observe that for uniform compound scaling, activations increase nearly \emph{linearly} with $s$.

\subsection{Complexity of Group Width Scaling}

\begin{table}\centering
\tablestyle{4.5pt}{1.00}
\begin{tabular}{@{}c|rr|rrr@{}}
 dims & \multicolumn{2}{c|}{scaling} & flops $(f)$ & params $(p)$ & acts $(a)$\\\shline
 none & $w$ & $g$ & $w g r^2$ & $w g$ & $w r^2$ \\
 $w$ & $\s w$ & $g$ & $\s w g r^2$ & $\s w g$ & $\s w r^2$ \\
 $g$ & $w$ & $\s g$ & $\s w g r^2$ & $\s w g$ & ${\color{scolor}1} w r^2$ \\
 $wg$ & $\ss w$ & $\ss g$ & $\s w g r^2$ & $\s w g$ & $\ss w r^2$ \\
\end{tabular}\vspace{1.5mm}
\caption{\textbf{Group width scaling}: Complexity of scaling a \emph{group} conv by scaling only width $w$, only group width $g$, or both (we assume $g \le w$). Scaling only $g$ does not impact activations; scaling uniformly in both $w$ and $g$ results in $\sqrt{s}$ increase in activations. Unless noted, we scale $g$ proportionally to $w$ for group conv.}
\label{tab:scaling:groups}
\end{table}

Many top-performing networks rely heavily on \emph{group conv} and \emph{depthwise conv}. A group conv with channel width $w$ and \emph{group width} $g$ is equivalent to splitting the $w$ channels into $w/g$ groups each of width $g$, applying a regular conv to each group, and concatenating the results. Depthwise conv is a special case with $g=1$. Therefore, its complexity is:
 \eqnsm{-1.5mm}{conv_g}{f = w g r^2, \quad p = wg, \quad a = w r^2.}
In Table~\ref{tab:scaling:groups} we show three basic strategies for scaling group conv. We observe that to obtain scaling behavior similar to scaling regular conv, both channel width and group width must be scaled. Therefore, unless otherwise noted, we scale $g$ proportionally to $w$. For networks that use depthwise conv ($g=1$), as in previous work~\cite{Tan2019}, we do not scale $g$.

Finally, we note that when scaling $g$, we must ensure $w$ is divisible by $g$. To address this, we set $g=w$ if $g > w$ and round $w$ to be divisible by $g$ otherwise ($w$ will change by at most 1/3 under such a strategy~\cite{Radosavovic2020}).

\section{Runtime of Scaled Models}\label{sec:runtime}

Our motivation is to design scaling strategies that result in \emph{fast} and \emph{accurate} models. In \S\ref{sec:complexity} we analyzed the behavior of flops, parameters, and activations for various scaling strategies. In this section we examine the relationship between these complexity metrics and model runtime. This will allow us to design new \emph{fast} scaling strategies in \S\ref{sec:fast}.

\begin{figure}[t]\centering
\includegraphics[height=2.38cm]{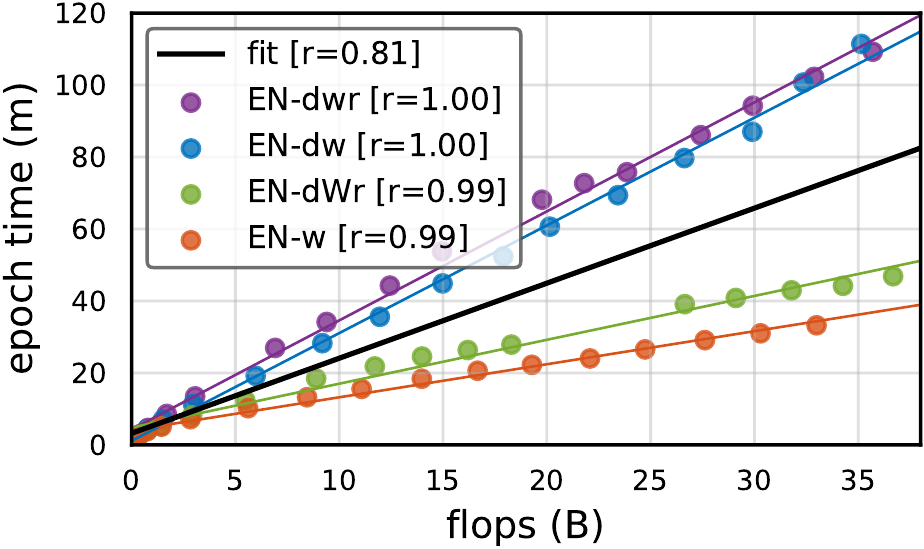}\hspace{2mm}
\includegraphics[height=2.38cm]{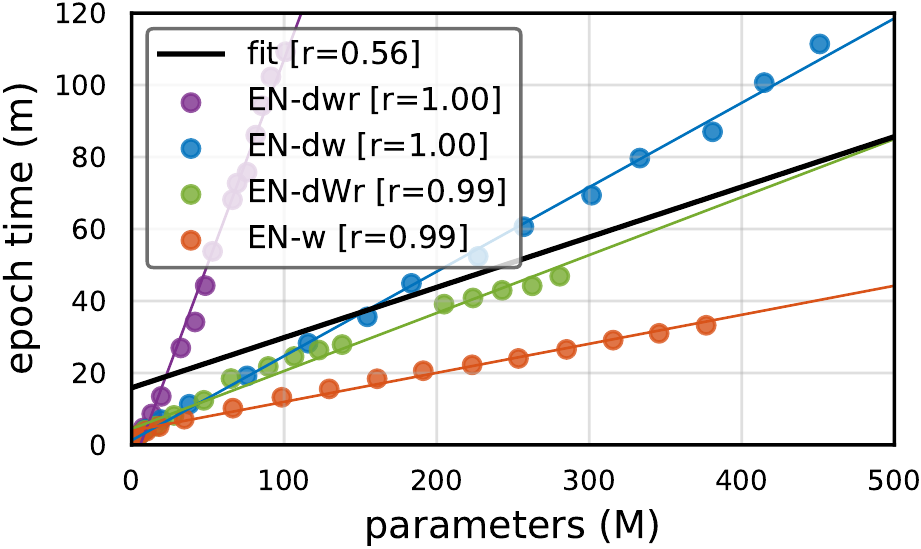}\\[1mm]
\includegraphics[height=2.38cm]{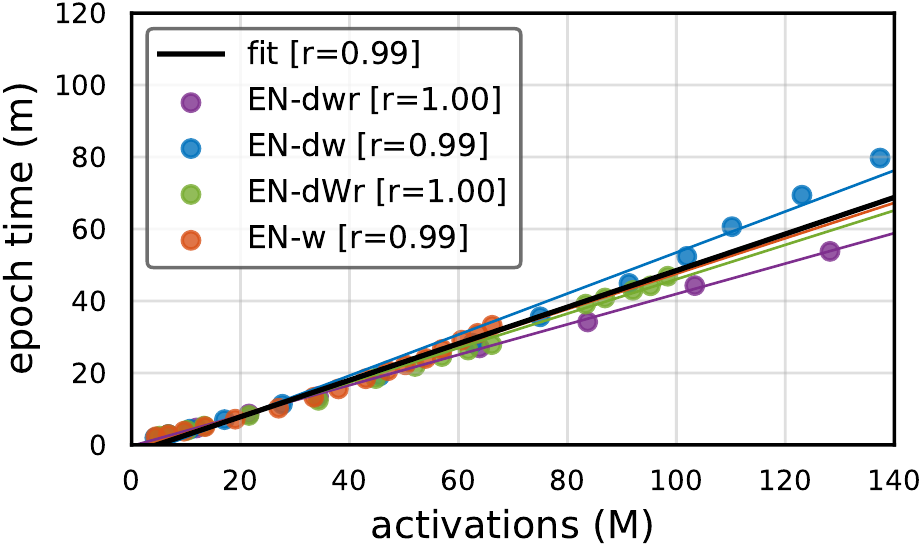}\hspace{2mm}
\includegraphics[height=2.38cm]{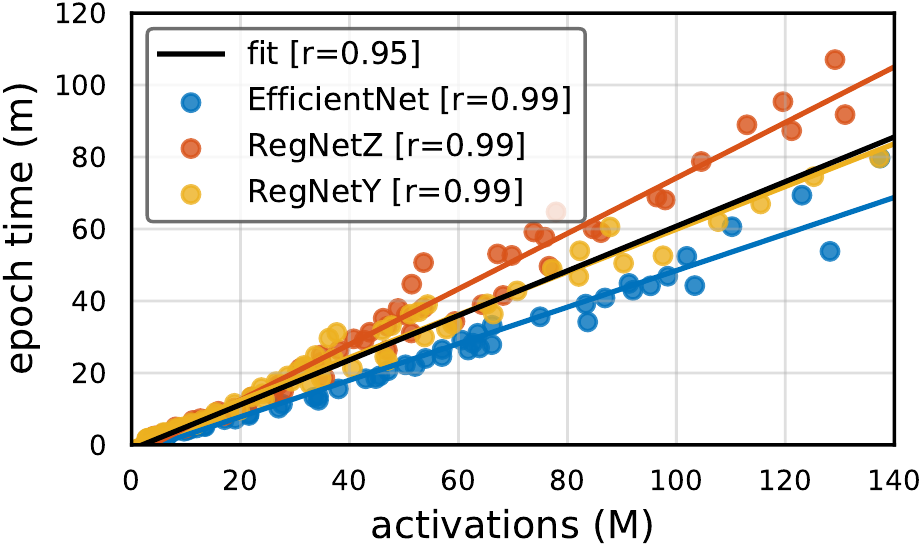}
\caption{\textbf{Model runtime} as a function of various complexity metrics. \emph{(Top-left):} We scale EfficientNet-B0 (EN-B0) using four scaling strategies ($dwr$, $dw$, $dWr$, $w$) with a wide range of scaling factors ($s<100$). For each scaling strategy we plot epoch time versus flops for each model (along with a best fit line). For a single scaling strategy (\eg, $w$), runtime is highly correlated with flops (\eg, Pearson's $r=0.99$). However, when comparing scaled versions of the same model using different scaling strategies, flops are only \emph{weakly} predictive of runtime ($r=0.81$). \emph{(Top-right):} Using the same set of models, we plot runtime versus parameters, and again observe parameters are even more weakly correlated with runtime ($r=0.56$). \emph{(Bottom-left):} Repeating the same analysis for runtime versus activations, we see that activations are \emph{strongly} predictive of runtime regardless of the scaling strategy ($r=0.99$). \emph{(Bottom-right):} We repeat the analysis of runtime versus activations for three models (see \S\ref{sec:exps:baselines} for model details). For scaled versions of each model, activations are highly predictive of runtime ($r \ge 0.99$), and only very large models tend to be flop bound. This makes activations an excellent proxy for runtime. We note, however, that activations are less predictive of runtime when comparing scaled versions of different models ($r=0.95$).}
\label{fig:timing}
\end{figure}

How are the complexity metrics we analyzed in \S\ref{sec:complexity} related to model runtime on modern accelerators? To answer this question, in Figure~\ref{fig:timing} we report runtime for a large number of models scaled from three base models as a function of flops, parameters, and activations. From these plots we can make two observations: flops and parameters are only \emph{weakly} predictive of runtime when scaling a single model via different scaling strategies; however, activations are \emph{strongly} predictive of runtime for a model regardless of the scaling strategy. See Figure~\ref{fig:timing} for additional details.

This simple result leads us to \emph{use model activations as a proxy for runtime}. Specifically, for scaled versions of a single model, the Pearson correlation between runtime and activations is $r \ge 0.99$, regardless of the scaling strategy, while correlation with flops and parameters is far lower ($r$ of 0.81 and 0.56, respectively). We caution, however, that activations cannot perfectly predict runtime across heterogeneous models ($r=0.95$), as models may use operations with different runtimes, \eg ReLU \vs SiLU. Moreover, some big models have runtimes higher than predicted from their activations indicating these models are \emph{flop bound}.

\paragraph{Implementation details.} We report the time to perform \emph{one epoch of training} on ImageNet~\cite{Deng2009} which contains \app1.2M training images. For each model, we use the largest batch size that fits in memory. We note that inference time is highly correlated with training time, but we report epoch time as it is easy to interpret (inference performance depends heavily on the use case). We time all models using PyTorch and 8 32GB Volta GPUs. Runtime is of course hardware dependent; however, we believe timing on GPUs is reasonable for two reasons. First, hardware accelerators (such as GPUs, TPUs, \etc.) are highly prevalent. Second, accelerators are extremely efficient in terms of compute but tend to be memory-bandwidth bound~\cite{Yang2019}, and this trend is expected to become more pronounced.

\section{Fast Compound Model Scaling}\label{sec:fast}

\begin{table}\centering
\tablestyle{5pt}{1.1}
\begin{tabular}{@{}c|c|ccc|ccc@{}}
 dims & $\alpha$ & $e_d$ & $e_w$ & $e_r$ & $f$ & $p$ & $a$\\\shline
 & $\alpha$ & $\frac{1-\alpha}{2}$ & $\alpha$ & $\frac{1-\alpha}{2}$
 & $\s$ & $\spow{\frac{1+\alpha}{2}}$ & $\spow{\frac{2-\alpha}{2}}$ \\
 $dr$ & 0 & 0.5 & 0 & 0.5 & $\s$ & $\spow{0.50}$ & $\spow{1.00}$ \\
 $dwr$ & 1/3 & 1/3 & 1/3 & 1/3 & $\s$ & $\spow{0.67}$ & $\spow{0.83}$ \\
 $dWr$ & 0.8 & 0.1 & 0.8 & 0.1 & $\s$ & $\spow{0.90}$ & $\spow{0.60}$ \\
 $w$ & 1 & 0 & 1 & 0 & $\s$ & $\spow{1.00}$ & $\spow{0.50}$ \\
\end{tabular}\vspace{1.5mm}
\caption{\textbf{Fast scaling.} We introduce a \emph{family} of scaling strategies parameterized by $\alpha$. The top row shows scaling factor exponents $e_d$, $e_w$, and $e_r$ as a function of $\alpha$, and the relative increase in model complexity as a function of $\alpha$ and scaling factor $s$. The remaining rows show instantiations of the scaling strategy for various $\alpha$. $\alpha=1$ corresponds to width ($w$) scaling, and $\alpha=1/3$ corresponds to uniform compound scaling ($dwr$). The new regime we explore in this work is $1/3<\alpha<1$. In particular, using $\alpha$ near 1 results in fewer activations and thus faster networks (see Figure~\ref{fig:timing}). In our experiments, we find $\alpha=0.8$ results in an excellent tradeoff between speed and accuracy. We use $dWr$ to denote fast scaling to emphasize scaling is primarily, but not only, along $w$.}
\label{tab:scaling:fast}
\end{table}

Given the strong dependency of runtime on activations, we aim to design scaling strategies that minimize the increase in model activations. As our results from Tables~\ref{tab:scaling:single}-\ref{tab:scaling:groups} indicate, of all scaling strategies that involve scaling width, depth, and resolution, scaling a network by increasing its \emph{channel width} and \emph{group width} results in the smallest increase in activations. Indeed, it is well known that \emph{wide} networks are quite efficient in wall-clock time~\cite{Zagoruyko2016}. Unfortunately, wide networks may not always achieve top results compared to deeper or higher-resolution models~\cite{He2016, Tan2019}.

To address this, in this work we introduce the concept of \emph{fast compound model scaling}, or simply \emph{fast scaling} for brevity. The idea is simple: we design and test scaling strategies that primary increase model width, but also increase depth and resolution to a lesser extent.

We formalize this by introducing a \emph{family} of scaling strategies parameterized by $\alpha$. Given $\alpha$ we define:
 \eqnsm{-1.5mm}{alpha1}{e_d = \tfrac{1-\alpha}{2}, \quad e_w = \alpha, \quad e_r=\tfrac{1-\alpha}{2},}
and when scaling a network by a factor of $s$, we set:
 \eqnsm{-1.5mm}{alpha2}{d' = s^{e_d} d, \quad w' = \sqrt{s}^{e_w} w, \quad r' = \sqrt{s}^{e_r} r.}
If using group conv, we also set $g'= \sqrt{s}^{e_w} g$ (same scaling as for $w$). The resulting complexity of the scaled model is:
 \eqnsm{-1.5mm}{alpha3}{f = \s d w^2 r^2, \quad p = \spow{\frac{1+\alpha}{2}} d w^2, \quad a=\spow{\frac{2-\alpha}{2}}d w r^2.}
Instantiations for scaling strategies using various $\alpha$ are shown in Table~\ref{tab:scaling:fast}. Setting $\alpha=1$ results in width ($w$) scaling (lowest activations). Setting $\alpha=0$ results in depth and resolution ($dr$) scaling (highest activations). $\alpha=1/3$ corresponds to uniform compound scaling ($dwr$).

The interesting new regime we explore is $1/3<\alpha<1$. In particular, we refer to scaling strategies with $\alpha$ near 1 as fast scaling. Unless specified, we use $\alpha=0.8$ by default, which we denote using $dWr$. Next, in \S\ref{sec:exps} we show that fast scaling results in good speed \emph{and} accuracy.

\section{Experiments}\label{sec:exps}

In this section we evaluate the effectiveness of our proposed fast scaling strategy. We introduce the baseline networks we test along with optimization settings in \S\ref{sec:exps:baselines}. In \S\ref{sec:exps:compound}, we evaluate existing scaling strategies, then we perform extensive experiments and comparisons of fast scaling in \S\ref{sec:exps:fast}. Finally we compare scaling \vs random search in \S\ref{sec:exps:sampling} and compare larger models in \S\ref{sec:exps:soa}.

\subsection{Baselines and Optimization Settings}\label{sec:exps:baselines}

\paragraph{Baseline networks.} In this work we evaluate scaling strategies on three networks families: EfficientNet~\cite{Tan2019}, RegNetY~\cite{Radosavovic2020}, and RegNetZ (described below). We chose these models as they are representative of the state-of-the-art and are well suited for our scaling experiments. Moreover, EfficientNet was introduced in the context of model scaling work~\cite{Tan2019}, making it an excellent candidate for our study. 

\paragraph{EfficientNet.} EfficientNets have been shown to achieve a good flop-to-accuracy tradeoff. These models use inverted bottlenecks~\cite{Sandler2018}, depthwise conv, and the SiLU nonlinearity~\cite{Hendrycks2016} (also popularly known as Swish~\cite{Ramachandran2017}). An EfficientNet is composed of seven stages with varying width, depth, stride and kernel size. The original model (EfficientNet-B0) was optimized in the mobile regime (400MF) using neural architecture search~\cite{Tan2019b} and scaled to larger sizes (B1-B7) via compound scaling. For further details, please see~\cite{Tan2019}.

Note that EfficientNets are specified by \app30 parameters (input resolution, 7 stages with 4 parameters each, and stem and head width). Given this high-dimensional search space, optimizing an EfficientNet is only feasible in a low-compute regime, and scaling must be used to obtain larger models.

\paragraph{RegNets.} As an alternative to neural architecture search, Radosavovic \etal~\cite{Radosavovic2020} introduced the idea of \emph{designing design spaces}, where a design space is a parameterized population of models. Using this methodology,~\cite{Radosavovic2020} designed a design space consisting of simple, regular networks called RegNets that are effective across a wide range of block types and flop regimes. Importantly for our work, a RegNet model is specified by a handful of parameters (\app 6), which then allows for fast model selection using random search. Thus, unlike EfficientNets, RegNets allow us to compare large models obtained either via \emph{scaling} or \emph{random search}.

\begin{figure}[t]\centering
\includegraphics[width=\linewidth]{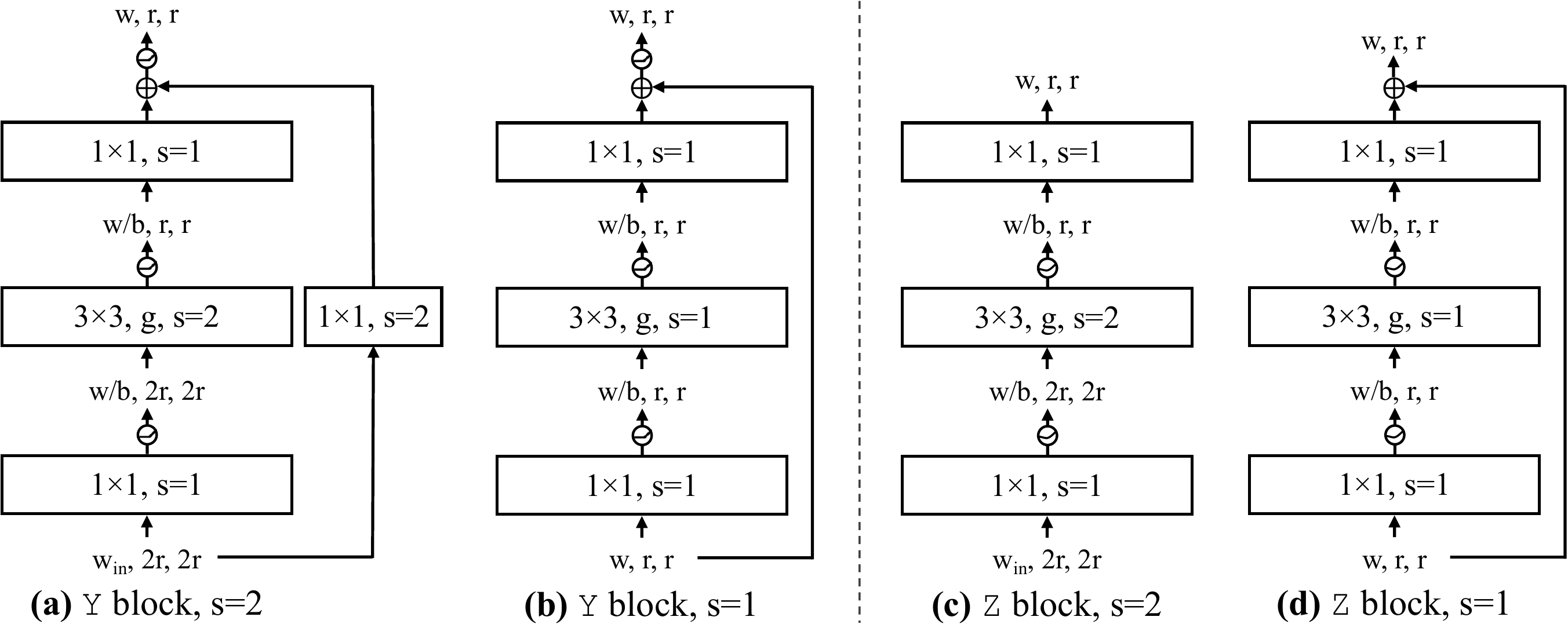}
\caption{\textbf{RegNet blocks.} Each stage consists of a stride $s=2$ block that halves $r$ and increases $w$ followed by multiple stride $s=1$ blocks with constant $r$ and $w$. (a-b) The Y block is based on \emph{residual bottlenecks} with group conv~\cite{Xie2017}. Each block consists of a $1\x1$ conv, a $3\x3$ group conv, and a final $1\x1$ conv. The $1\x1$ convs can change $w$ via the bottleneck ratio $b$, however, we set $b=1$ following~\cite{Radosavovic2020}. BatchNorm~\cite{Ioffe2015} and ReLU follow each conv. (c-d) We introduce the Z block based on \emph{inverted bottlenecks}~\cite{Sandler2018}. The Z block is similar to the Y block with 4 differences: no non-linearity follows the final $1\x1$ conv, (2) SiLU~\cite{Hendrycks2016} is used in place of ReLU, (3) the stride 2 variant of the block has no residual, and (4) $b<1$ (we use $b=1/4$ in all experiments). Finally, a Squeeze-and-Excitation (SE) op~\cite{Hu2018} (reduction ratio of $1/4$) follows the $3\x3$ conv for both the Y and Z blocks (not shown).}
\label{fig:regnet}
\end{figure}

A RegNet consists of a stem, a body with four stages, and a head. Each stage consists of a sequence of identical blocks. The block type can vary depending on the model (the two block types we use are shown in Figure~\ref{fig:regnet}). Importantly, the widths and depths of a RegNet are not specified independently per stage, but are determined by a \emph{quantized linear function} which has 4 parameters ($d$, $w_0$, $w_a$, $w_m$), for details see~\cite{Radosavovic2020}. Any other block parameters (like group width or bottleneck ratio) are kept constant across stages.

\paragraph{RegNetY.} The RegNetY block (Y) is shown in Figure~\ref{fig:regnet}~(a-b). The Y block resembles the standard \emph{residual bottleneck} block with group conv~\cite{Xie2017}. Additionally it uses a Squeeze-and-Excitation (SE) layer~\cite{Hu2018}. Following~\cite{Radosavovic2020}, we set the bottleneck ratio $b$ to 1 (effectively no bottleneck). A RegNetY model is thus fully specified with 5 parameters: $d$, $w_0$, $w_a$, $w_m$, and $g$. Unlike~\cite{Radosavovic2020}, we additionally vary the image input resolution $r$ (bringing the total parameters to 6).

\paragraph{RegNetZ.} We introduce a new Z block based on \emph{inverted bottlenecks}~\cite{Sandler2018}. The Z block resembles the Y block except it omits the last nonlinearity and inverts the bottleneck (we use $b=1/4$ in all experiments). See Figure~\ref{fig:regnet}~(c-d) for additional details. A RegNetZ model, built using the Z block, is fully specified with the same 6 parameters as a RegNetY model. We note that EfficientNet also uses inverted bottlenecks, but we introduce RegNetZ to allow us to compare large models obtained via scaling \emph{and} random search.

\paragraph{Optimization settings.} Our goal is to enable \emph{fair} and \emph{reproducible} results. However, we also aim to achieve \emph{state-of-the-art} results. This creates a tension between using a simple yet weak optimization setup (\eg,~\cite{Radosavovic2020}) versus a strong setup that yields good results but may be difficult to reproduce (\eg,~\cite{Tan2019}). To address this, we use a training setup that effectively balances between these two objectives.

Our setup is as follows: we use SGD with a momentum of 0.9, label smoothing with $\epsilon=0.1$~\cite{Szegedy2016a}, mixup with $\alpha=0.2$~\cite{Zhang2018mixup}, AutoAugment~\cite{Cubuk2018}, stochastic weight averaging (SWA)~\cite{Dai2020}, and mixed precision training~\cite{Micikevicius2018}. For all models we use 5 epochs of gradual warmup~\cite{Goyal2017}. We use an exponential learning rate schedule with a batch size of 1024 (distributed on 8 32GB GPUs), learning rate $\lambda=2.0$, and decay $\beta=0.02$.\footnote{We parameterize the exponential learning rate via $\lambda_t = \lambda \beta^\frac{t}{T}$, where $t$ is the current epoch, $T$ the final epoch, $\lambda$ is the initial learning rate, and $\lambda\beta$ is the final learning rate. We use this parameterization (as opposed to $\lambda_t = \lambda \gamma^t$) as it allows us to use a single setting for the decay $\beta$ regardless of the schedule length $T$ (setting $\gamma=\beta^{1/T}$ makes the two equivalent).} For RegNets we use a weight decay of 2e-5 and for EfficientNets we use 1e-5. Batch norm parameters are not decayed. For large models we reduce the batch size and learning rate proportionally as in~\cite{Goyal2017}. For reproducibility, we will release code for our setup.

\begin{table}[t]\centering
\resizebox{\columnwidth}{!}{\tablestyle{4pt}{1.05}
\begin{tabular}{@{}l|cccc|cc|ccc@{}}
  & flops & params & acts & time & \multicolumn{2}{c|}{publication} & \multicolumn{3}{c}{schedule} \\
  & (B)   & (M)    & (M)  & (min) & ICML & arXiv & 1$\x$ & 2$\x$ & 4$\x$ \\\shline
 EN-B0 & 0.4 & 5.3 & 6.7 & 2.8 & 23.7 & 22.7 & 23.6\mypm{0.09} & 22.7\mypm{0.08} & \bf 22.3\mypm{0.04} \\
 EN-B1 & 0.7 & 7.8 & 10.9 & 4.6 & 21.2 & 20.8 & 21.7\mypm{0.18} & 20.8\mypm{0.10} & \bf20.5\mypm{0.09} \\
 EN-B2 & 1.0 & 9.1 & 13.8 & 5.9 & 20.2 & 19.7 & 20.7\mypm{0.06} & 20.0\mypm{0.12} & \bf19.6\mypm{0.09} \\
 EN-B3 & 1.8 & 12.2 & 23.8 & 9.5 & 18.9 & \bf 18.3 & 19.4\mypm{0.07} & 18.8\mypm{0.10} & \bf18.3\mypm{0.11} \\
 EN-B4 & 4.4 & 19.3 & 49.5 & 19.2 & 17.4 & \bf 17.0 & 18.0\mypm{0.05} & 17.4\mypm{0.07} & 17.3\mypm{0.06} \\
 EN-B5 & 10.3 & 30.4 & 98.9 & 40.8 & 16.7 & \bf 16.3 & 17.1\mypm{0.13} & 16.7\mypm{0.05} & -- \\
\end{tabular}}\vspace{1.5mm}
\caption{\textbf{EfficientNet reproduction.} The first set of results includes the originally reported errors (from ICML~\cite{Tan2019} and updated numbers later reported on arXiv), the second set our reproduction under three schedule lengths (1$\x$ corresponds to 100 epochs), averaged over 3 trials. Our results match or outperform the originally reported results (for the biggest nets they slightly lag the updated arXiv errors). We emphasize that unlike the results in~\cite{Tan2019}, we use the same, easy to reproduce optimization setup for all models.}
\label{tab:baselines:en}
\end{table}

\begin{table}[t]\centering
\resizebox{\columnwidth}{!}{\tablestyle{4pt}{1.05}
\begin{tabular}{@{}l|cccc|ccc@{}}
  & flops & params & acts & time  & \multicolumn{3}{c}{schedule} \\
  & (B)   & (M)    & (M)  & (min) & 1$\x$ & 2$\x$ & 4$\x$ \\\shline
 RegNetY-500MF & 0.5 & 5.6 & 4.2 & 2.3 & 24.8\mypm{0.07} & 23.9\mypm{0.14} & 23.2\mypm{0.05} \\
 RegNetZ-500MF & 0.5 & 7.1 & 5.9 & 3.1 & 22.2\mypm{0.04} & 21.3\mypm{0.02} & 21.0\mypm{0.08} \\\hline
 RegNetY-4GF-224 & 4.0 & 20.6 & 12.3 & 6.4 & 19.4\mypm{0.07} & 18.4\mypm{0.05} & 18.1\mypm{0.07} \\
 RegNetY-4GF & 4.1 & 22.4 & 14.5 & 7.7 & 18.8\mypm{0.04} & 18.0\mypm{0.07} & 17.7\mypm{0.09} \\
 RegNetZ-4GF-224 & 4.0 & 26.9 & 20.8 & 11.3 & 17.7\mypm{0.04} & 17.2\mypm{0.06} & 17.0\mypm{0.04} \\
 RegNetZ-4GF & 4.0 & 28.1 & 24.3 & 11.7 & 17.5\mypm{0.09} & 17.0\mypm{0.12} & 16.9\mypm{0.04} \\
\end{tabular}}\vspace{1.5mm}
\caption{\textbf{RegNet Baselines}. The two 500MF RegNet models use a default resolution of 224. RegNetY-4GF-224 uses the default 224 resolution; RegNetY-4GF uses a resolution found by random search. Likewise there are two versions of RegNetZ-4GF with default and discovered resolutions.}
\label{tab:baselines:regnet}
\end{table}

\paragraph{EfficientNet baselines.} In Table~\ref{tab:baselines:en}, we report EfficientNet results using our optimization setup versus results from~\cite{Tan2019}. We report our results using a `1$\x$', `2$\x$', or `4$\x$' schedule (corresponding to 100, 200, and 400 epochs, respectively). Our 2$\x$ schedule achieves competitive results, our 4$\x$ schedule outperforms the originally reported results for all but the largest model tested. We use the 2$\x$ schedule in all following experiments unless otherwise noted.

\paragraph{RegNet baselines.} In Table~\ref{tab:baselines:regnet} we report results for baseline RegNet models. We obtain these models via random search as in~\cite{Radosavovic2020}.\footnote{We sample RegNet model configurations until we obtain 32 models in a given flop regime, train each of these model using the 1$\x$ schedule, and finally select the best one. Sampling just 32 random models in a given flop regime is typically sufficient to obtain accurate models as shown in~\cite{Radosavovic2020}.} Note that there are two versions of the 4GF RegNets (using default and discovered resolutions).

\begin{figure}[t]\centering
\includegraphics[height=2.4cm]{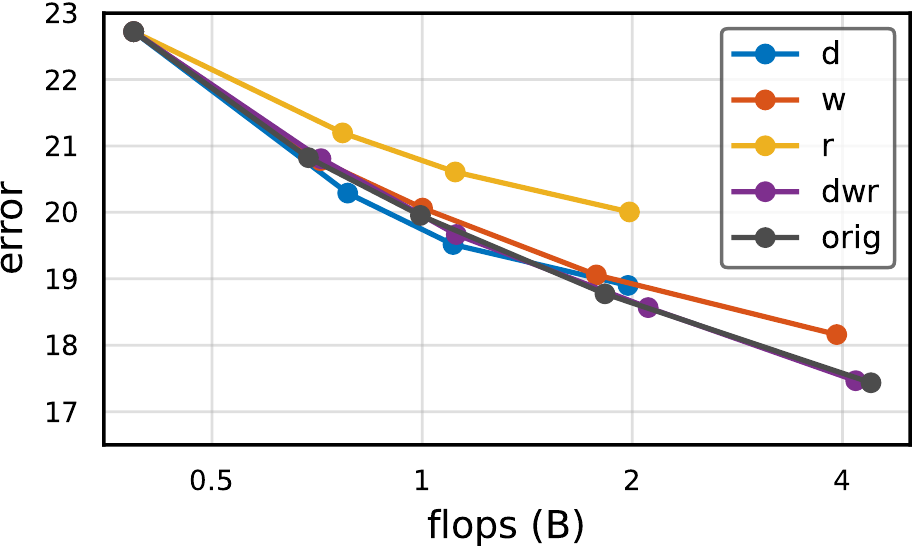}\hspace{2mm}
\includegraphics[height=2.4cm]{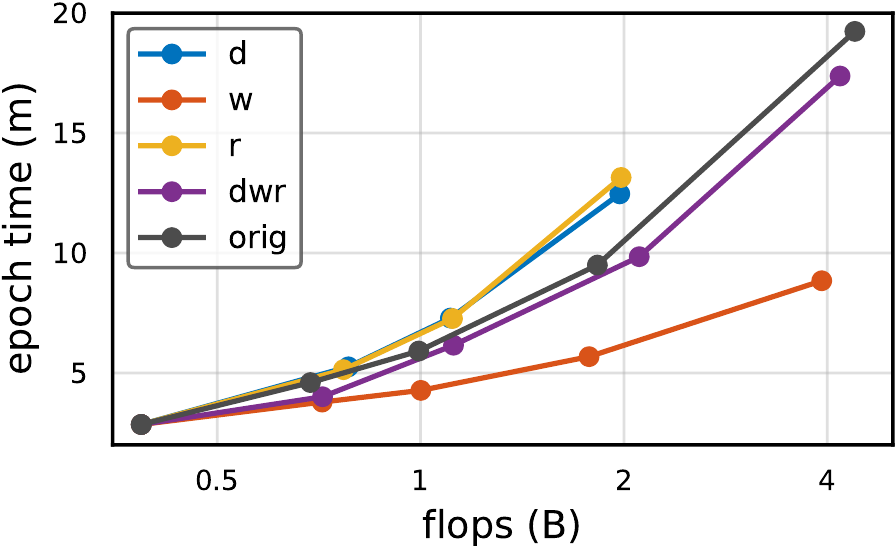}\\
\caption{\textbf{Compound scaling: EfficientNet.} (Left) Uniform compound scaling ($dwr$) offers the best accuracy relative to simple scaling along depth ($d$), width ($w$), or resolution ($r$). All models are scaled from EfficientNet-B0 (400MF) up to at most 4GF. (Right) Models obtained with $w$ scaling are much faster than those from $dwr$ scaling. Both of these results are expected. However, as we will show, it is possible to obtain models that are both fast \emph{and} accurate. For reference, we also show the original EfficientNet models (orig) obtained via non-uniform compound scaling~\cite{Tan2019}, the results closely match uniform compound scaling ($dwr$).}
\label{fig:exp:compound:effnet}
\end{figure}

\begin{figure}[t]\centering
\includegraphics[height=2.4cm]{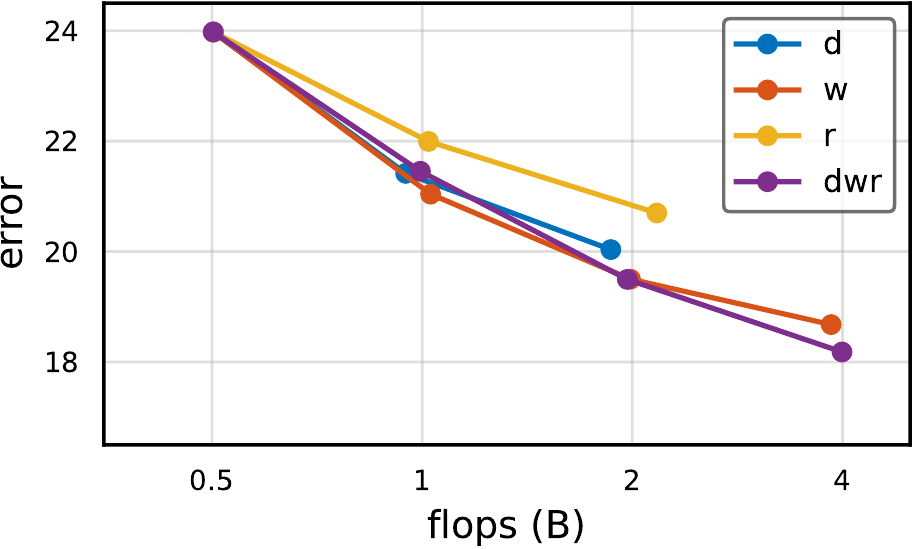}\hspace{2mm}
\includegraphics[height=2.4cm]{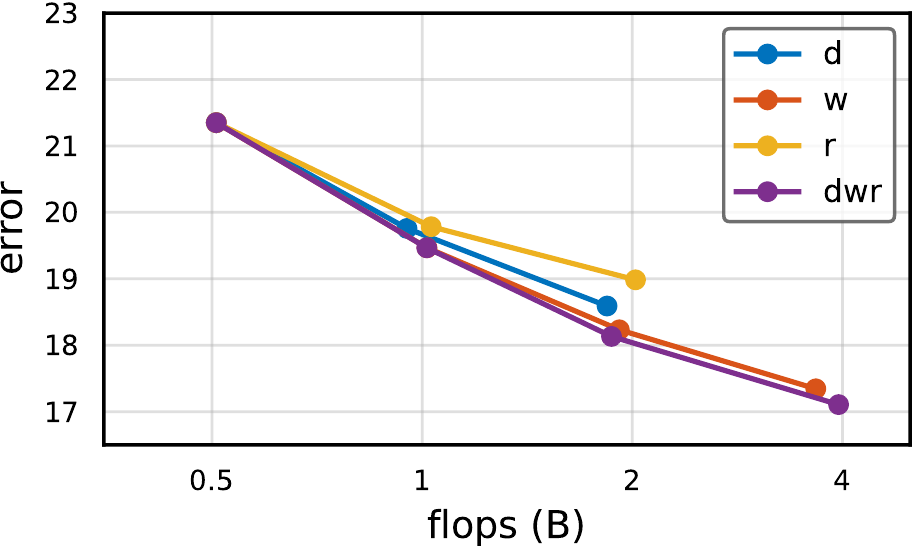}\\
\caption{\textbf{Compound scaling: RegNet.} We apply simple and compound scaling to RegNetY-500MF (left) and RegNetZ-500MF (right). As in Figure~\ref{fig:exp:compound:effnet}, $dwr$ scaling achieves the best error, but at significant increase in runtime (see appendix) relative to $w$ scaling.}
\label{fig:exp:compound:regnet}
\end{figure}

\subsection{Simple and Compound Scaling}\label{sec:exps:compound}

We now turn to evaluation of simple and compound scaling~\cite{Tan2019} described in \S\ref{sec:complexity:simple} and \S\ref{sec:complexity:compound}, respectively. For these experiments we scale the baseline models from \S\ref{sec:exps:baselines}.

In Figure~\ref{fig:exp:compound:effnet}, we evaluate the accuracy (left) and runtime (right) of EfficientNet-B0 scaled either via simple scaling along width ($w$), depth ($d$), or resolution ($r$) or via uniform compound scaling ($dwr$). As expected, $dwr$ scaling provides the best accuracy, but results in slower models than $w$ scaling. This suggests a tradeoff between speed and accuracy, but as we will show shortly, this need not be the case. Finally, we tested uniform scaling along pairs of dimensions (see Table~\ref{tab:scaling:compound}), but $dwr$ scaling proved best (not shown).

We also compare uniform compound scaling ($dwr$) to the original compound scaling rule (orig) from~\cite{Tan2019}, which empirically set the per-dimension scalings factors. As expected from our analysis in \S\ref{sec:complexity:compound}, $dwr$ scaling is close in both accuracy and runtime to the original compound scaling rule without the need to optimize individual scaling factors.

In Figure~\ref{fig:exp:compound:regnet} we repeat the same experiment but for the RegNetY-500MF and RegNetZ-500MF baselines. We see a similar behavior, where $dwr$ scaling achieves the strongest results. Runtimes (see appendix) exhibit very similar behaviors ($w$ scaling is much faster). Note that as discussed, group width $g$ is scaled proportionally to width $w$.

\begin{figure}[t]\centering
\includegraphics[height=2.4cm]{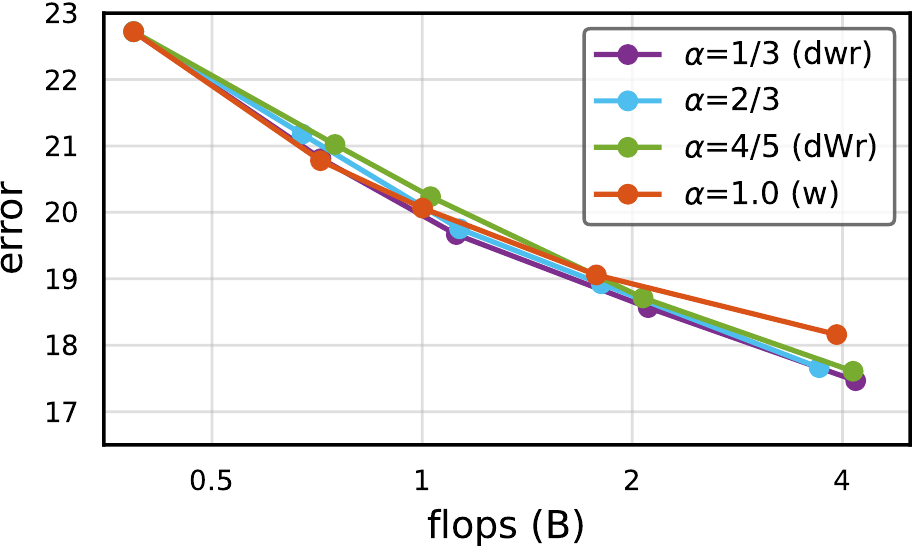}\hspace{2mm}
\includegraphics[height=2.4cm]{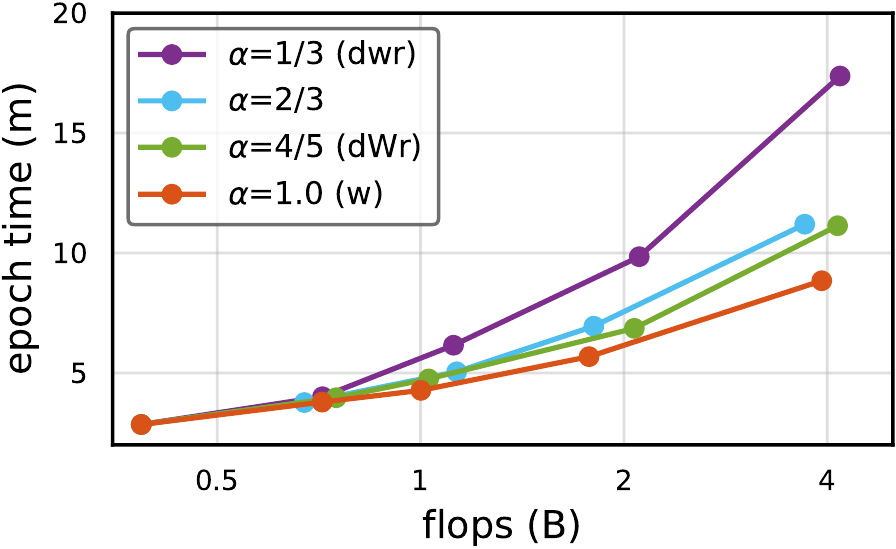}\\
\caption{\textbf{Fast scaling: EfficientNet.} We test scaling EfficientNet-B0 using our family of scaling strategies parameterized by $\alpha$ (see Table~\ref{tab:scaling:fast}). (Left) Scaling with any $\alpha<1$ achieves good accuracy and results in a sizable gap in error to scaling with $\alpha=1$ ($w$). The exact value of $\alpha<1$ does not greatly influence the error. (Right) While all scaling strategies with $\alpha<1$ give good accuracy, their runtime differ substantially. A setting of $\alpha=4/5$ ($dWr$) gives the best of both worlds: models that are both fast \emph{and} accurate.}
\label{fig:exp:fast:effnet}
\end{figure}

\begin{figure}[t]\centering
\includegraphics[height=2.4cm]{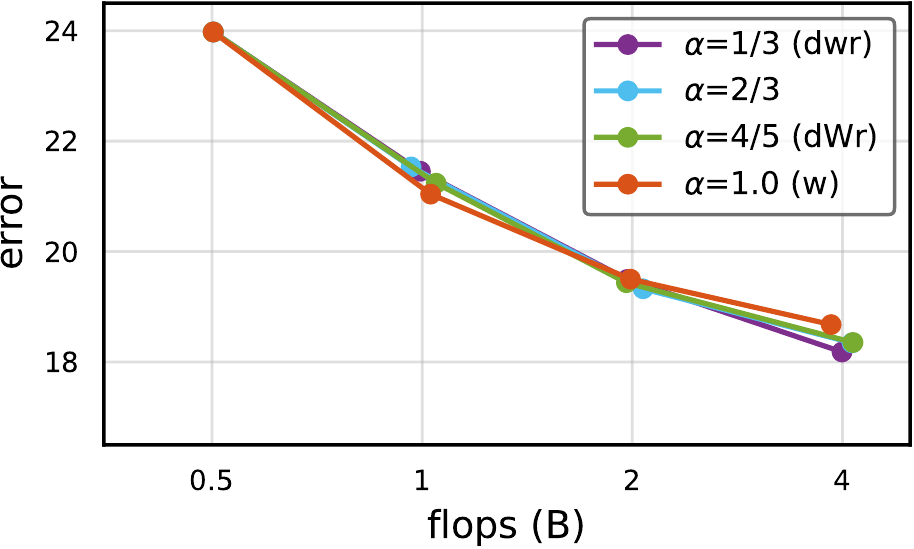}\hspace{2mm}
\includegraphics[height=2.4cm]{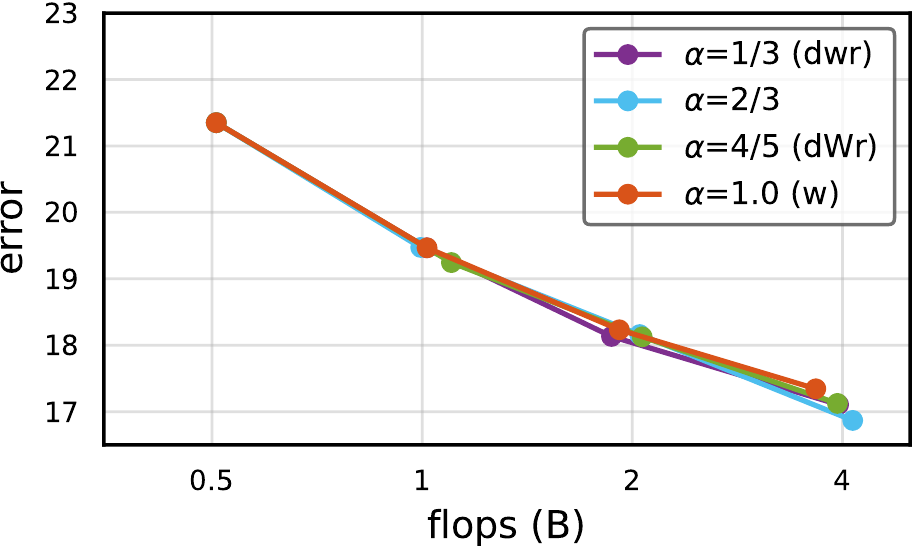}\\
\caption{\textbf{Fast scaling: RegNet.} We apply scaling with different $\alpha$ to RegNetY-500MF (left) and RegNetZ-500MF (right). As in Figure~\ref{fig:exp:fast:effnet}, $dWr$ scaling yields good accuracy \emph{and} speed (see appendix for rutnimes). We note that $\alpha$ may potentially be be further tuned to tradeoff speed and accuracy, but we use $\alpha=4/5$ in this work.}
\label{fig:exp:fast:regnet}\vspace{-1mm}
\end{figure}

\subsection{Fast Scaling}\label{sec:exps:fast}

We now perform an empirical analysis of the \emph{effectiveness} of our fast scaling strategy. Recall that in \S\ref{sec:fast} we introduced a family of scaling strategies parameterized by $\alpha$ that interpolates between uniform compound scaling ($dwr$) when $\alpha=1/3$ to width scaling ($w$) when $\alpha=1$. As $\alpha$ goes toward 1, the model activations increase least as we scale a model, resulting in faster models. In particular, we define $\alpha=4/5$ as \emph{fast scaling}, and denote it by $dWr$.

In Figure~\ref{fig:exp:fast:effnet}, we evaluate the accuracy (left) and runtime (right) of EfficientNet-B0 scaled with various settings of $\alpha$. Interestingly, for all tested values of $\alpha<1$ model accuracy was quite similar and substantially higher than for $w$ scaling ($\alpha=1$), especially for larger models. In terms of runtime, $dWr$ scaling is nearly as fast as $w$ scaling, and substantially faster than $dwr$ scaling. We emphasize that the differences in memory and speed increase \emph{asymptotically}, hence the difference in runtime for models scaled with different $\alpha$ becomes more pronounced at larger scales. 

In Figure~\ref{fig:exp:fast:regnet} we repeat the same experiment but for the RegNet baselines. Results are similar, $dWr$ scaling ($\alpha=4/5$) achieves excellent accuracy and runtime. Finally, we observe that for RegNets, $w$ scaling is more effective than for EfficientNet. This can be partially explained as for RegNets we scale the group width $g$ along width $w$ (EfficientNet always uses $g=1$), indeed setting $g=1$ and scaling RegNets by just $w$ performs worse (see appendix).

\subsection{Scaling versus Search}\label{sec:exps:sampling}

\begin{table}[t]\centering\vspace{-1mm}
\resizebox{\columnwidth}{!}{\tablestyle{4pt}{1.05}
\begin{tabular}{@{}l|cccc|ccc@{}}
  & flops & params & acts & time  & \multicolumn{3}{c}{schedule} \\
  & (B)   & (M)    & (M)  & (min) & 1$\x$ & 2$\x$ & 4$\x$ \\\shline
 RegNetY-500MF$\rightarrow$4GF & 4.1 & 36.2 & 13.3 & 7.2 & 19.1\mypm{0.07} & 18.6\mypm{0.09} & 18.3\mypm{0.06} \\
 RegNetY-4GF [optimized] & 4.1 & 22.4 & 14.5 & 7.7 & 18.8\mypm{0.04} & 18.0\mypm{0.07} & 17.7\mypm{0.09} \\\hline
 RegNetZ-500MF$\rightarrow$4GF & 4.0 & 41.1 & 19.4 & 10.5 & 17.7\mypm{0.07} & 17.2\mypm{0.07} & 17.0\mypm{0.05} \\
 RegNetZ-4GF [optimized] & 4.0 & 28.1 & 24.3 & 11.7 & 17.5\mypm{0.09} & 17.0\mypm{0.12} & 16.9\mypm{0.04} \\\hline
 RegNetY-500MF$\rightarrow$16GF & 16.2 & 112.7 & 29.4 & 17.8 & 17.8\mypm{0.18} & 17.2\mypm{0.06} & 16.9\mypm{0.10} \\
 RegNetY-4GF$\rightarrow$16GF & 15.5 & 72.3 & 30.7 & 16.4 & 17.3\mypm{0.09} & 16.8\mypm{0.11} & 16.6\mypm{0.03} \\
\end{tabular}}\vspace{1.5mm}
\caption{\textbf{Scaling \vs Search}. Models optimized for a given flop regime (via random search) outperform scaled models (rows 1-4). Nevertheless, scaling is necessary in flop regimes where optimization is computationally prohibitive. A \emph{hybrid} approach is to optimize a model in an intermediate regime (\eg 4GF) prior to scaling to a higher flop regime (\eg 16GF), as in rows 5-6.}
\label{tab:sampling}
\end{table}

\begin{figure}[t]\centering
\includegraphics[height=2.4cm]{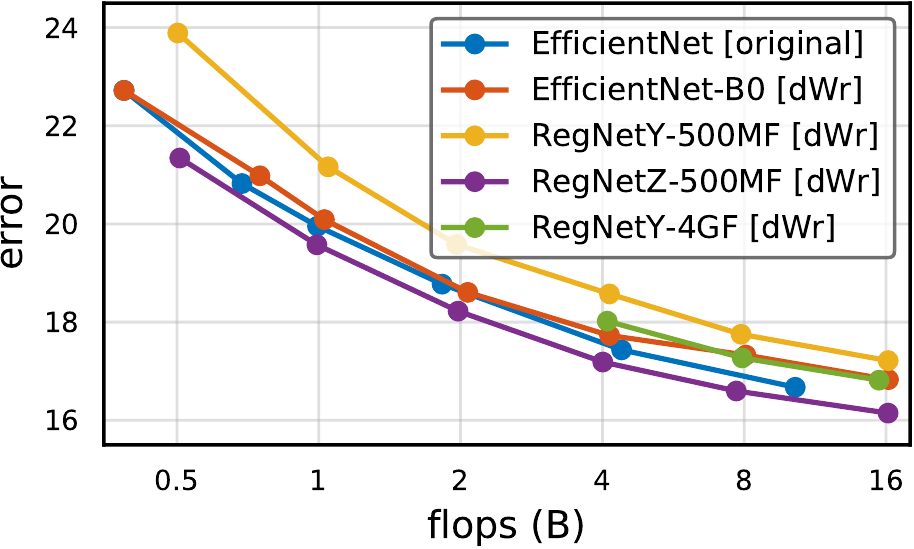}\hspace{2mm}
\includegraphics[height=2.4cm]{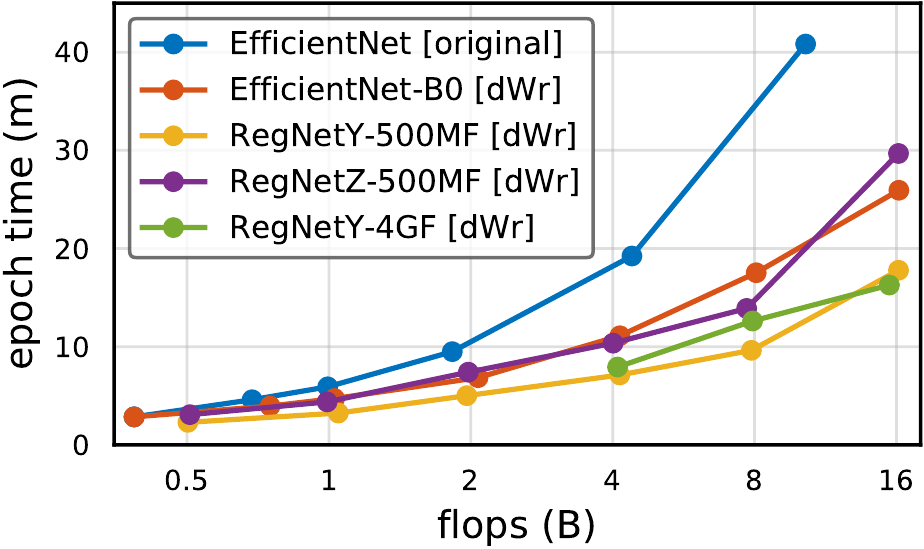}
\caption{\textbf{Large models.} We scale four models via fast scaling ($dWr$) up to 16GF (1$\x$ to 32$\x$ scaling). We include the original EfficientNet model for reference. All results use our 2$\x$ schedule. See \S\ref{sec:exps:soa} for details and discussion.}
\label{fig:exp:soa}
\end{figure}

How do \emph{scaled} models compare to models obtained via \emph{random search}? Recall that RegNets only have 6 free parameters, so optimizing a RegNet directly by random search in an intermediate flop regime is feasible (see \S\ref{sec:exps:baselines}).

Table~\ref{tab:sampling} compares three sets of models. First, we compare RegNetY at 4GF obtained either via $dWr$ scaling (denoted by RegNetY-500MF$\rightarrow$4GF) or search (RegNetY-4GF) in rows 1-2. The best sampled model outperforms the scaled model by 0.6\% with a 4$\x$ schedule. We repeat this analysis for RegNetZ (rows 3-4) and find the best sampled model outperforms the scaled model by 0.1\%. These results indicate that scaling a high-accuracy model is not guaranteed to yield an optimal model. Nevertheless, scaling is often necessary for targeting high compute regimes where model optimization is not feasible.

The above results suggest a \emph{hybrid scaling strategy}, in which we optimize a model at an intermediate flop regime prior to scaling the model to larger scales. In Table~\ref{tab:sampling}, rows 5-6, we compare two 16GF RegNetY models, one scaled by 32$\x$ from a 500MF model and one scaled 4$\x$ from an optimized 4GF model. The model obtained with the hybrid strategy of scaling an intermediate model is 0.3\% better. 

Finally, observe that the best sampled models have far fewer parameters than the scaled models. We found that at higher flop regimes, optimized models have fewer blocks in the last stage, which greatly reduces their parameters. This shows a limitation of uniformly scaling model stages without redistributing blocks across stages.

\begin{table}[t]\centering\vspace{-1mm}
\resizebox{\columnwidth}{!}{\tablestyle{4pt}{1.05}
\begin{tabular}{@{}l|cccc|ccc@{}}
  & flops & params & acts & time  & \multicolumn{3}{c}{schedule} \\
  & (B)   & (M)    & (M)  & (min) & 1$\x$ & 2$\x$ & 4$\x$ \\\shline
 ResNet50~\cite{He2016} & 4.1 & 25.6 & 11.3 & 3.5 & 22.0\mypm{0.12} & 21.0\mypm{0.08} & 20.5\mypm{0.07} \\
 ResNeXt50~\cite{Xie2017} & 4.2 & 25.0 & 14.6 & 5.8 & 20.8\mypm{0.06} & 19.9\mypm{0.16} & 19.5\mypm{0.05} \\
 EfficientNet-B4~\cite{Tan2019} & 4.4 & 19.3 & 49.5 & 19.2 & 18.0\mypm{0.05} & 17.4\mypm{0.07} & 17.3\mypm{0.06} \\ 
 RegNetY-4GF & 4.1 & 22.4 & 14.5 & 7.7 & 18.8\mypm{0.04} & 18.0\mypm{0.07} & 17.7\mypm{0.09} \\
 RegNetZ-4GF & 4.0 & 28.1 & 24.3 & 11.7 & 17.5\mypm{0.09} & 17.0\mypm{0.12} & 16.9\mypm{0.04} \\\hline
 EfficientNet-B0$\rightarrow$4GF & 4.1 & 36.1 & 29.2 & 11.1 & 18.4\mypm{0.11} & 17.7\mypm{0.07} & 17.4\mypm{0.11} \\
 RegNetY-500MF$\rightarrow$4GF & 4.1 & 36.2 & 13.3 & 7.2 & 19.1\mypm{0.07} & 18.6\mypm{0.09} & 18.3\mypm{0.06} \\
 RegNetZ-500MF$\rightarrow$4GF & 4.0 & 41.1 & 19.4 & 10.5 & 17.7\mypm{0.07} & 17.2\mypm{0.07} & 17.0\mypm{0.05} \\\hline
 EfficientNet-B0$\rightarrow$16GF & 16.2 & 122.8 & 61.8 & 25.8 & 17.4\mypm{0.08} & 16.8\mypm{0.09} & --\\
 RegNetY-500MF$\rightarrow$16GF & 16.2 & 112.7 & 29.4 & 17.8 & 17.8\mypm{0.18} & 17.2\mypm{0.06} & 16.9\mypm{0.10} \\
 RegNetY-4GF$\rightarrow$16GF & 15.5 & 72.3 & 30.7 & 16.4 & 17.3\mypm{0.09} & 16.8\mypm{0.11} & 16.6\mypm{0.03} \\
 RegNetZ-500MF$\rightarrow$16GF & 16.2 & 134.8 & 42.6 & 29.4 & 16.6\mypm{0.04} & 16.1\mypm{0.06} & 16.1\mypm{0.07} \\
 RegNetZ-4GF$\rightarrow$16GF & 15.9 & 95.3 & 51.3 & 33.2 & 16.5\mypm{0.05} & 16.0\mypm{0.10} & 16.0\mypm{0.05} \\
\end{tabular}}\vspace{1.5mm}
\caption{\textbf{Large models.} For reference and reproducibility, we list details of our scaled 4GF and 16GF models models trained using our 1$\x$, 2$\x$, and 4$\x$ schedules. For reference, we also retrain ResNet50~\cite{He2016} and ResNeXt50~\cite{Xie2017} using our strong setup (and obtain an \app3\% reduced error than originally reported).}
\label{tab:soa}\vspace{-2mm}
\end{table}

\subsection{Comparison of Large Models}\label{sec:exps:soa}\vspace{-1mm}

The primary benefit of model scaling is it allows us to scale to larger models where optimization is not feasible. In Figure~\ref{fig:exp:soa}, we scale four models up to 16GF using fast scaling. We make the following observations:
\begin{enumerate}[topsep=2.5pt, itemsep=-0.8ex]
\item Model ranking is \emph{consistent} across flop regimes, with scaled versions RegNetZ achieving the best accuracy.
\item All models obtained via fast scaling ($dWr$) are asymptotically faster than the original EfficientNet models, including our scaled versions of EfficientNet-B0.
\item The \emph{gap} between the highest and lowest error models (RegNetY and RegNetZ) \emph{shrinks} from 2.2\% at 500MF to 0.8\% at 16GF, implying that on ImageNet model optimization may be less important at high flop regimes.
\item The \emph{hybrid} approach of scaling an intermediate flop regime model to higher flops (4GF$\rightarrow$16GF) closes much of the gap between RegNetY and RegNetZ.
\item RegNetY is the fastest model tested and a good choice if runtime is constrained, especially at higher flops.
\end{enumerate}

In Table~\ref{tab:soa} we give further details of the 4GF and 16GF models we tested, along with additional baselines. We note that RegNetY-4GF$\rightarrow$16GF uses less memory and is \emph{faster} than EfficientNet-B4, even though this RegNetY model has \app4$\x$ as many flops. This emphasizes the importance of looking at metrics beyond flops when comparing models.

\section{Discussion}\label{sec:discussion}\vspace{-1mm}

In this work we presented a general framework for analyzing model scaling strategies that takes into account not just flops but also other network properties, including activations, which we showed are highly correlated with runtime on modern hardware. Given our analysis, we presented a fast scaling strategy that primarily, but not exclusively, scales model width. Fast scaling results in accurate models that also have fast runtime. While the optimal scaling approach may be task dependent, we hope our work provides a general framework for reasoning about model scaling.

\section*{Appendix}

\begin{figure}[t]\centering
\includegraphics[height=2.38cm]{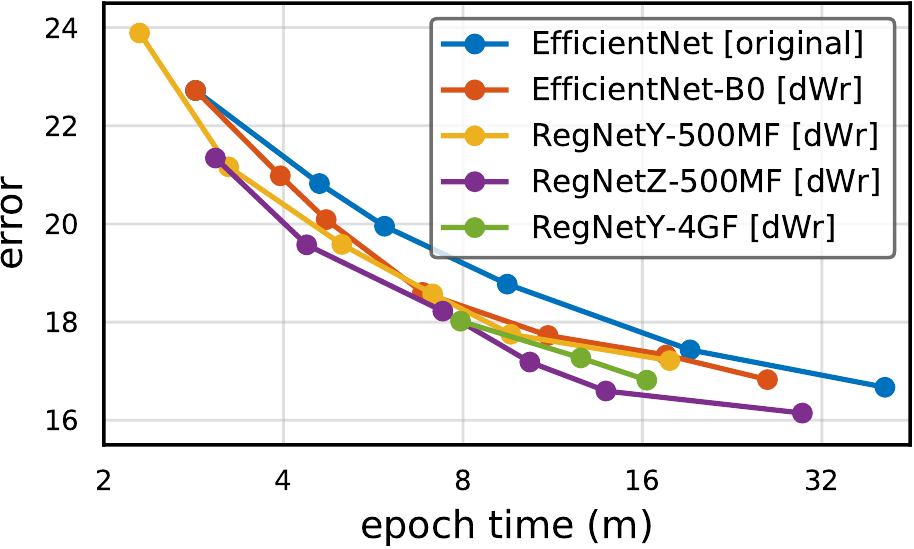}\hspace{2mm}
\includegraphics[height=2.38cm]{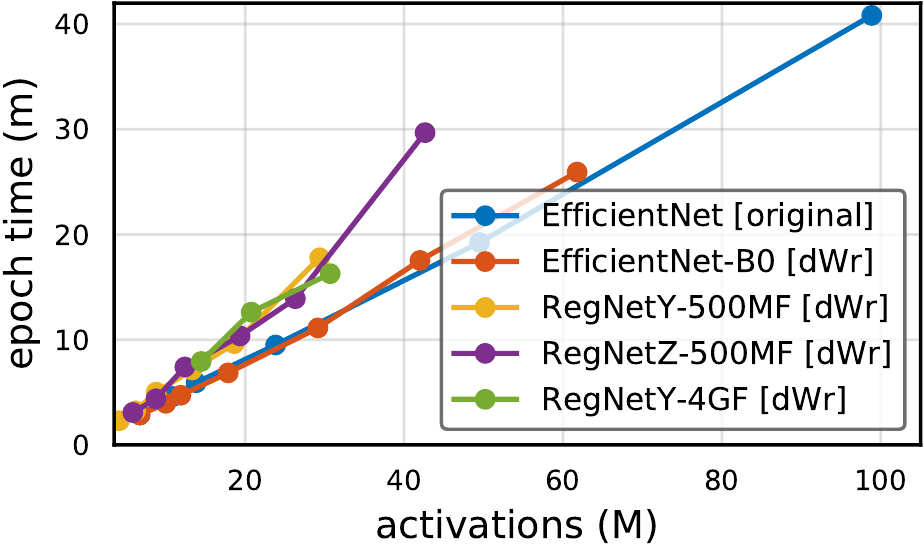}\\
\caption{\textbf{Large Models Additional Analysis} (see also Figure~\ref{fig:exp:soa}). \emph{(Left):} Plotting error versus runtime shows that scaled versions of RegNetY and RegNetZ offer the best speed versus accuracy tradeoff. However, the exact speed of these models is implementation dependent and may change with additional optimizations. \emph{(Right):} For a given model type, activations of these large state-of-the-art models are strongly predictive of runtime, as expected.}
\label{fig:soa:more}
\end{figure}

\paragraph{Large models additional analysis.} In Figure~\ref{fig:soa:more} we show further analysis of the models from  Figure~\ref{fig:exp:soa}. The left plot shows error versus runtime, with RegNetY and RegNetZ offering the best speed versus accuracy tradeoff. While offering a useful perspective, the relative ranking of methods is implementation dependent and may change with additional optimizations. For example, group conv seems to be underoptimized relative to depthwise or full-width conv, so a better implementation could lead to speedups for models that rely on group conv. On the other hand, activations are highly predictive of runtime of a scaled model (right plot), which we expect to hold regardless of implementation.

\begin{figure}[t]\centering
\includegraphics[height=2.38cm]{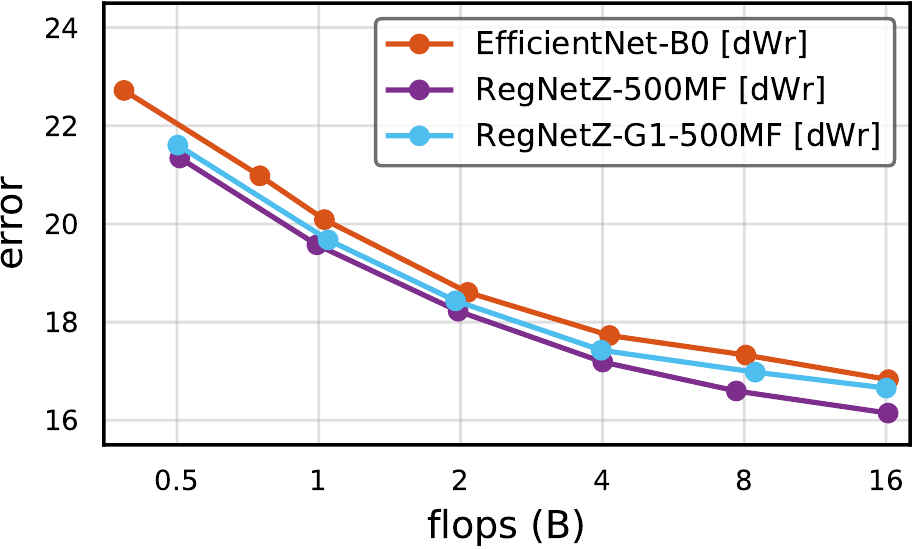}\hspace{2mm}
\includegraphics[height=2.38cm]{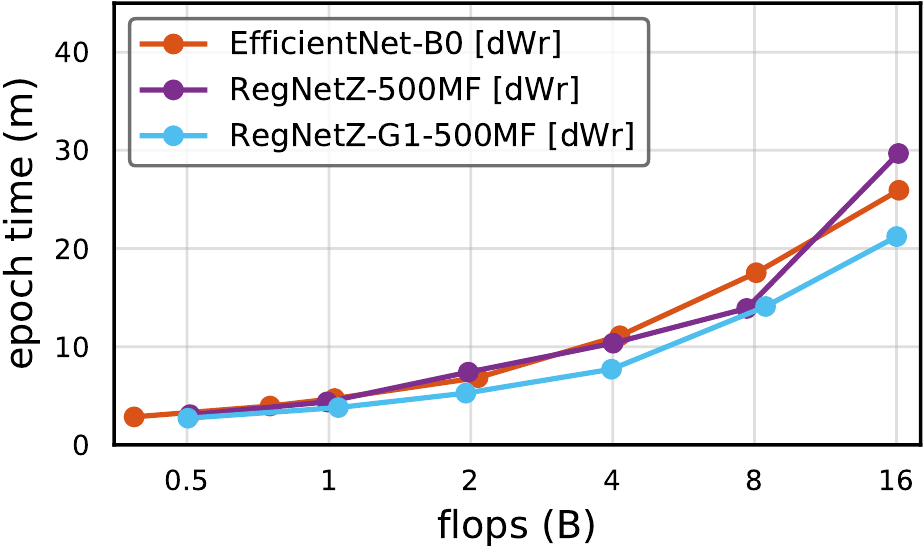}\\
\caption{\textbf{Group \vs Depthwise Conv.} EfficientNet uses depthwise conv while RegNetZ uses group conv, but otherwise the models use fairly similar components (inverted bottlenecks, SiLU, SE). To study this difference, we introduce RegNetZ-G1 which is like RegNetZ but uses depthwise conv. At higher flops, RegNetZ shows gains over RegNetZ-G1 and EfficientNet, demonstrating that group conv may be a better option at higher compute regimes.}
\label{fig:soa:groups}\vspace{-3mm}
\end{figure}

\paragraph{Group \vs depthwise conv.} EfficientNet~\cite{Liu2018} uses depthwise conv while RegNetZ uses group conv. Does this explain the accuracy difference between them? To answer this, we introduce a variant RegNetZ which is constrained to use depthwise conv, denoted as RegNetZ-G1. In Figure~\ref{fig:soa:groups} we plot scaled versions of EfficientNet-B0, RegNetZ-500MF, and RegNetZ-G1-500MF (using $dWr$ scaling). Interestingly, RegNetZ-G1 achieves better accuracy then EfficientNet, which is surprising as they use similar components and EfficientNet-B0 was obtained with a more sophisticated search. Nevertheless, we see that indeed much of the improvement of RegNetZ over EfficientNet, especially at higher flops, comes from using group conv.

\begin{figure}[t]\centering
\includegraphics[height=2.34cm]{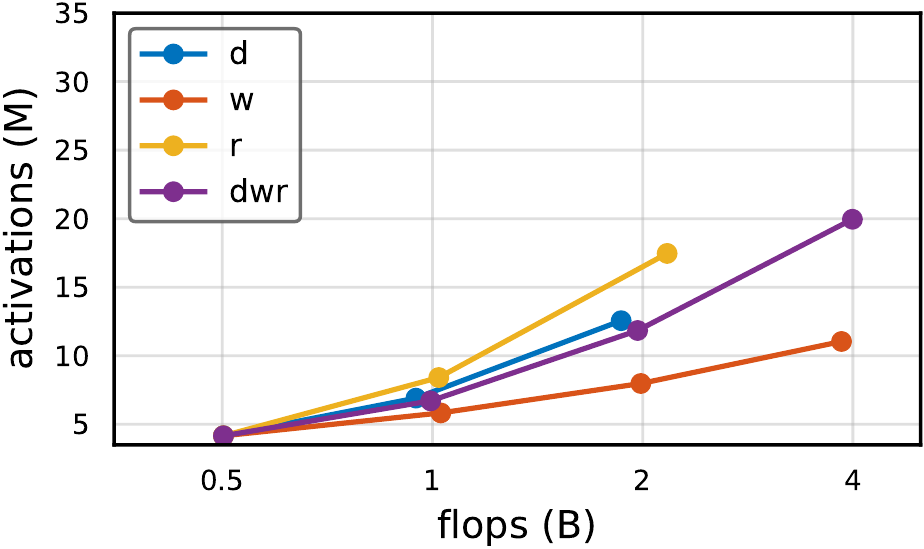}\hspace{2mm}
\includegraphics[height=2.34cm]{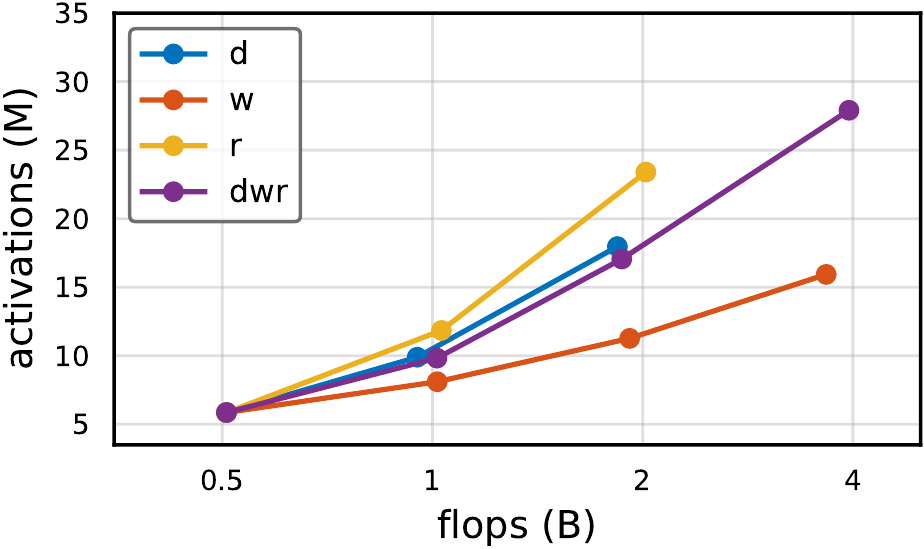}\\
\includegraphics[height=2.4cm]{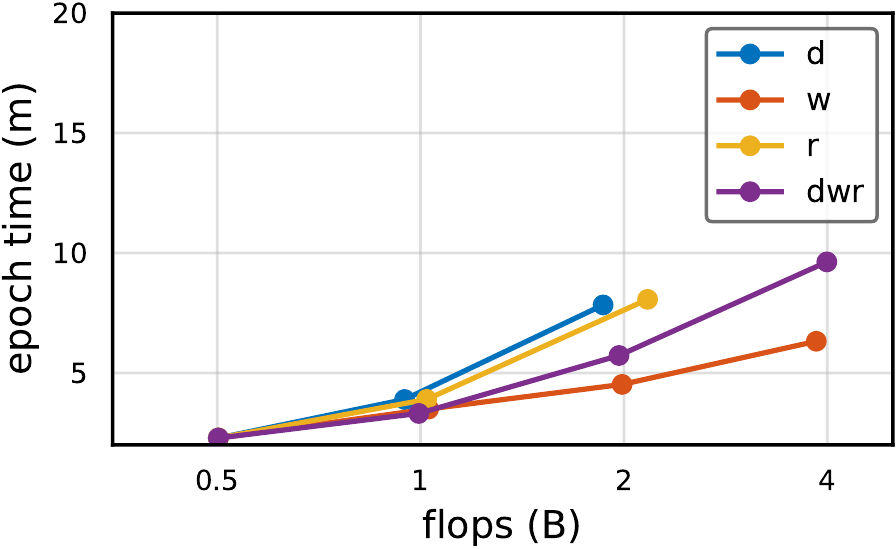}\hspace{2mm}
\includegraphics[height=2.4cm]{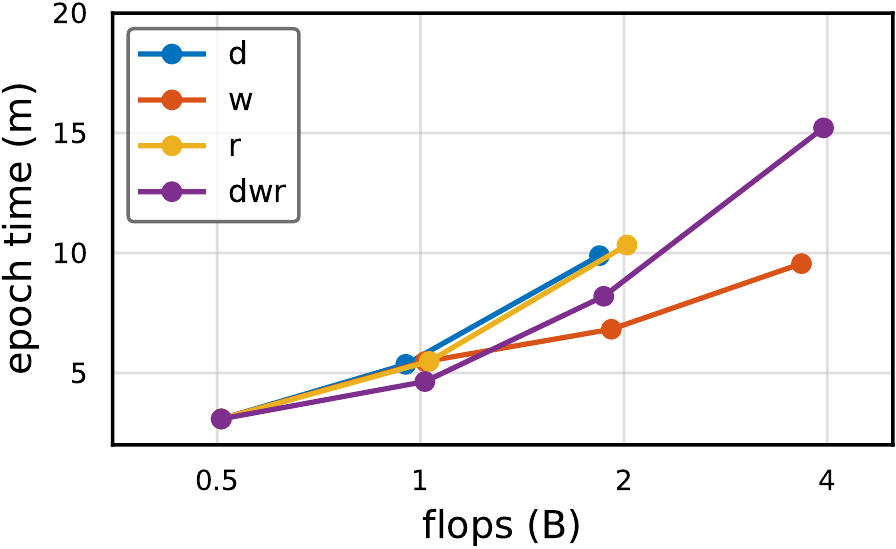}\\
\caption{\textbf{Compound scaling: RegNet.} Activations (top) and runtime (bottom) versus flops for the RegNetY (left) and RegNetZ (right) scaled models from Figure~\ref{fig:exp:compound:regnet}; shown here for completeness. Results are as expected, with activations being highly predictive of runtime and with $w$ scaling resulting in the fastest scaled models.}
\label{fig:exp:compound:regnet:time}
\end{figure}

\begin{figure}[t]\centering
\includegraphics[height=2.34cm]{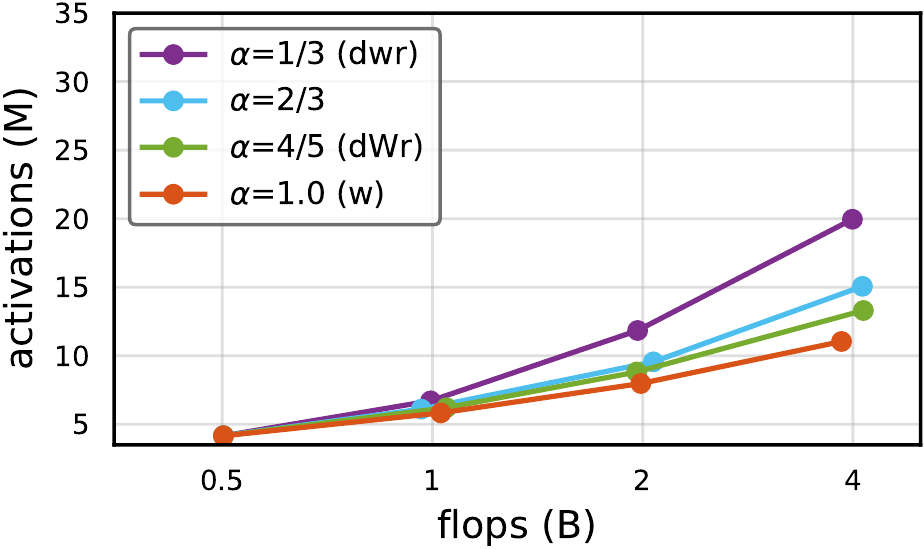}\hspace{2mm}
\includegraphics[height=2.34cm]{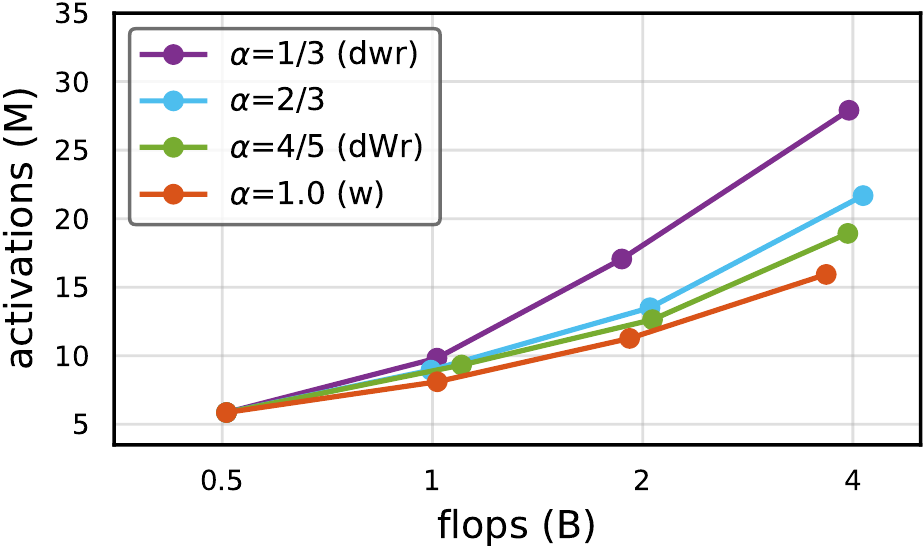}\\
\includegraphics[height=2.4cm]{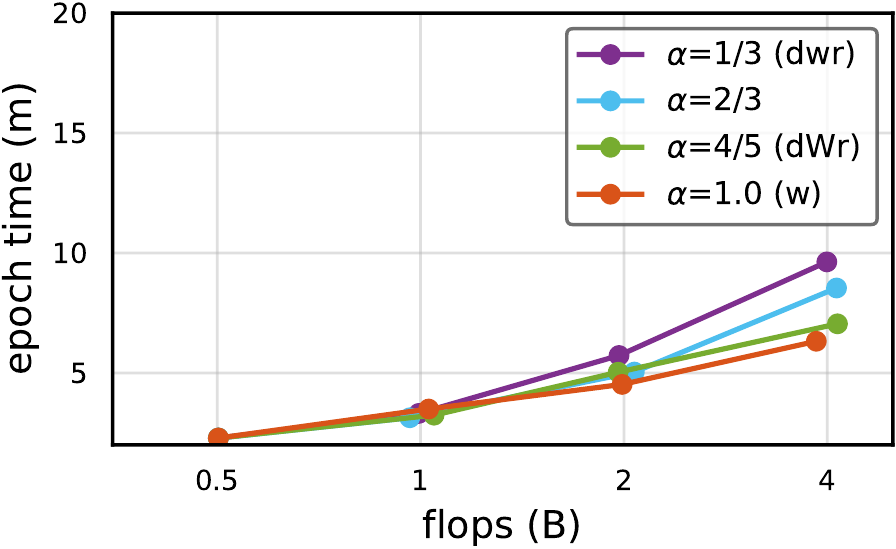}\hspace{2mm}
\includegraphics[height=2.4cm]{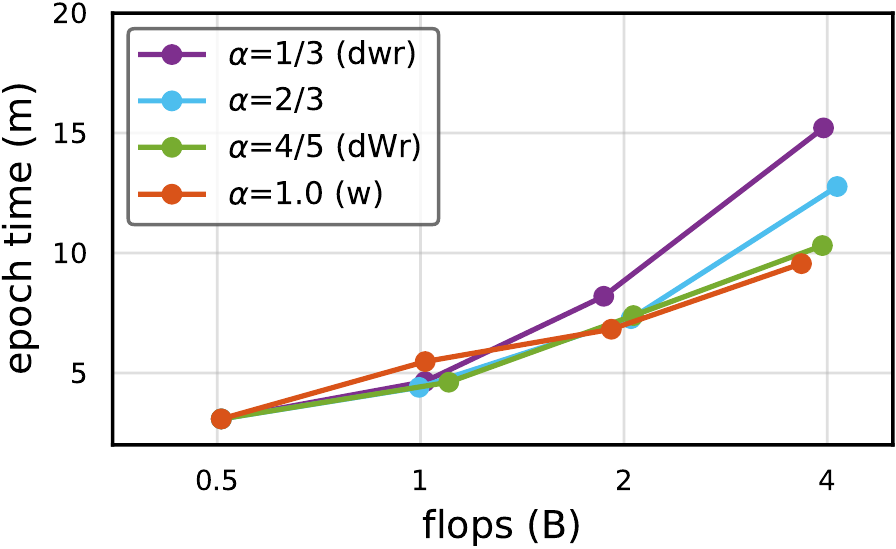}\\
\caption{\textbf{Fast scaling: RegNet.} Activations (top) and runtime (bottom) versus flops for the RegNetY (left) and RegNetZ (right) scaled models from Figure~\ref{fig:exp:fast:regnet}; shown here for completeness. Results are as expected, with activations being highly predictive of runtime and with large $\alpha$ resulting in the fastest scaled models.}
\label{fig:exp:fast:regnet:time}
\end{figure}

\paragraph{RegNet timing results.} For completeness, activation and runtime results for RegNetY and RegNetZ corresponding to the scaling strategies from Figure~\ref{fig:exp:compound:regnet} and \ref{fig:exp:fast:regnet} are shown in Figure~\ref{fig:exp:compound:regnet:time} and \ref{fig:exp:fast:regnet:time}, respectively. In both figures, activations are shown at the top and timings at the bottom, and RegNetY is shown on the left and RegNetZ on the right. First, observe that the model timing plots closely follow the model activations plots in all cases. This is expected since activations and timings are highly correlated (see \S\ref{sec:runtime}). Second, as expected, in Figure~\ref{fig:exp:compound:regnet:time} we see $w$ scaling results in lowest activations/runtime, and in Figure~\ref{fig:exp:fast:regnet:time} we see that using a large $\alpha$ results in lowest activations/runtime for all models.

\section*{Acknowledgements}
{\small We would like to thank Xiaoliang Dai for help with the simple yet strong training setup used in this work and Kaiming He and Ilija Radosavovic for valuable discussions and feedback.}

{\setstretch{.95}\small\bibliographystyle{ieee}\bibliography{scaling}}

\end{document}